\documentclass[11pt]{article}

\usepackage[final]{acl}

\usepackage{times}
\usepackage{latexsym}

\usepackage[T1]{fontenc}

\usepackage[utf8]{inputenc}

\usepackage{microtype}

\usepackage{inconsolata}

\usepackage{graphicx}

\newcommand{\brain}{\raisebox{-0.2ex}{\includegraphics[height=1em]{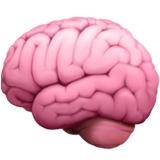}}}
\newcommand{\txt}{\raisebox{-0.2ex}{\includegraphics[height=1em]{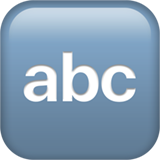}}}
\newcommand{\img}{\raisebox{-0.2ex}{\includegraphics[height=1em]{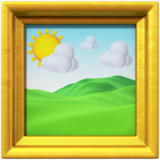}}}

\usepackage{amsmath}
\usepackage{booktabs}
\usepackage[table]{xcolor}
\usepackage{multirow}

\usepackage{amssymb}    
\usepackage{amsmath}    
\usepackage{diagbox}
\usepackage{tcolorbox}
\usepackage{enumitem}

\usepackage{xcolor}
\usepackage{listings}

\lstdefinestyle{promptstyle}{
    basicstyle=\ttfamily\small,
    breaklines=true,
    breakatwhitespace=true,
    breakindent=0pt,
    breakautoindent=false,
    postbreak={},
    frame=single,
    backgroundcolor=\color{gray!5},
    columns=fullflexible,
    keepspaces=false,
    showstringspaces=false,
    xleftmargin=0pt,
    xrightmargin=0pt,
    framesep=4pt,
    framexleftmargin=4pt,
    framexrightmargin=4pt
}

\newcommand{\posL}{\cellcolor{blue!8}}
\newcommand{\posM}{\cellcolor{blue!16}}
\newcommand{\posH}{\cellcolor{blue!26}}
\newcommand{\posVH}{\cellcolor{blue!38}}

\newcommand{\negL}{\cellcolor{red!8}}
\newcommand{\negM}{\cellcolor{red!16}}
\newcommand{\negH}{\cellcolor{red!26}}
\newcommand{\negVH}{\cellcolor{red!38}}

\title{Same Semantics, Different Outcome: On the Modality Robustness of Multimodal LLMs under Knowledge Conflict}

\author{
Jungyeon Lee, Yejin Yoon, Taeuk Kim\textsuperscript{*} \\
Hanyang University, Seoul, Republic of Korea \\
{\tt \{jungyune,stillwithyou,kimtaeuk\}@hanyang.ac.kr}
}

\begin{document}
\maketitle
\begin{abstract}
Multimodal large language models (MLLMs) are increasingly provided with contextual evidence in heterogeneous forms: as a text passage, as a rendered image of the same passage, or as both together.
However, it remains unclear how consistently these surface forms are processed, especially when the evidence conflicts with the model's parametric knowledge.
We study modality robustness under knowledge conflict across 13 MLLMs and two datasets, and find them far from robust. (1) Contrary to common belief, models favor a context that contradicts parametric knowledge more readily in image form than in text form; (2) when a contradicting text and image are presented together, the preferred modality is essentially arbitrary, varying with input order, model, and dataset.
We further demonstrate that this instability has practical consequences: it degrades performance in multimodal RAG and can be exploited by adversarial attacks.
To alleviate this brittleness, we examine several simple techniques---prompting, steering, supervised fine-tuning (SFT), and direct preference optimization; the majority prove ineffective, whereas SFT achieves moderate success.
We therefore call for greater awareness of this inconsistency and argue that it is fundamental, demanding attention at multiple training stages.
\end{abstract}

\renewcommand{\thefootnote}{} 
\footnotetext{\textsuperscript{*}Corresponding author.}
\renewcommand{\thefootnote}{\arabic{footnote}} 

\section{Introduction}
\label{sec:intro}

Multimodal large language models (MLLMs) are increasingly deployed in retrieval-augmented and agentic workflows, where evidence arrives in a variety of surface forms, e.g., plain text, webpages, PDFs, screenshots, or scanned documents \cite{yu2025visrag, faysse2025colpali}.
These workflows presuppose that MLLMs process semantically equivalent content consistently, irrespective of its modality.
That expectation is not guaranteed to hold, however, because inputs in different modalities typically follow distinct internal pathways through these models, beginning with modality-specific tokenization.



The risk posed by such latent inconsistency is greatly amplified under \textit{knowledge conflict} (KC), where external evidence contradicts the model's parametric knowledge or where multiple sources support competing answers.
In such scenarios, the answer may diverge drastically, or even reverse, according to which source the model accepts as authoritative; if that acceptance is governed by the form in which the evidence is delivered rather than by its content, the robustness and reliability of the model are severely compromised.

Although previous work \cite{deng2025words,sim-etal-2025-vlms,zhang2025evaluating} has reported that MLLMs exhibit a modality bias---typically placing more weight on textual than visual information---this claim has not been examined in knowledge-conflict scenarios.
Moreover, the few studies on multimodal knowledge conflict \cite{hua2025vision,nguyen2025challenges,zhang2025modalities,zhangrobust} focus mostly on natural images rather than documents, even though knowledge is typically organized in document form, not as scenes or faces.

To fill this gap, we investigate the \textit{modality robustness} of MLLMs through the lens of knowledge conflict.
We evaluate two setups: (1) a single piece of external evidence is delivered as text or as a rendered image (i.e., text-as-image); (2) two textual and visual contexts conflict with each other.
Unlike prior work, we also guarantee that parametric knowledge is distinct from any contextual knowledge, enabling more tightly controlled analysis.

Spanning multiple MLLMs, datasets, and conflict categories, the dominant pattern we observe is \textit{instability}: which modality a model trusts shifts unpredictably with input order, dataset construction, and model family, and no single factor reliably explains the variation.
Crucially, this contradicts the widely reported textual bias in MLLMs. 
In our single-evidence setting, the direction is in fact reversed: when the image is a rendering of the same text, MLLMs tend to prefer it over the text itself. 
Moreover, even this reversed preference dissolves once both modalities are presented in conflict, a setting that prior work has not thoroughly examined.


We further examine whether this sensitivity has practical consequences beyond controlled configurations, using two downstream tasks.
First, in multimodal RAG, we keep a passage's content fixed and vary only its format---raw text or rendered image---to test whether that change alone alters how models use retrieved evidence (\S\ref{subsec:rag}).
We find that MLLMs are less distracted by irrelevant context when salient knowledge is represented as an image rather than text, suggesting a remedy for the lost-in-the-middle phenomenon \cite{liu-etal-2024-lost}.

Second, we examine safety-critical settings, in which the surface form of a harmful request may affect the model's refusal behavior (\S\ref{subsec:safety}).
Our results indicate that rendering harmful requests as images increases attack success rates by 6.99 percentage points on average, underscoring a serious and readily exploitable vulnerability.

Finally, we explore methods to improve the modality robustness of MLLMs.
We observe that simple prompting is insufficient; instead, fine-tuning on knowledge-conflict data partially mitigates the problem, although it does not completely eliminate the inconsistency (\S\Ref{sec:mitigation}).
Consequently, we call for greater community awareness of MLLMs' inconsistency across input modalities, and argue that this problem is fundamental, requiring further investigation at different stages of model training.


To summarize, our contributions are as follows:
\begin{itemize}[itemsep=0pt, leftmargin=10pt]
    \item We formalize modality robustness as an evidence-authority problem under knowledge conflict.
    \item We construct single- and multi-evidence conflict settings that disentangle semantic content, presentation modality, and parametric knowledge.
    \item Across 13 MLLMs and two datasets, modality reliance varies across model families, input orders, and evidence compositions, with downstream effects on multimodal RAG and refusal behavior.
    \item Comparing several mitigation strategies, we find conflict-aware fine-tuning to be the most effective at balancing modality preference.
\end{itemize}

\begin{figure*}[t]
\centering
\includegraphics[width=0.95\linewidth]{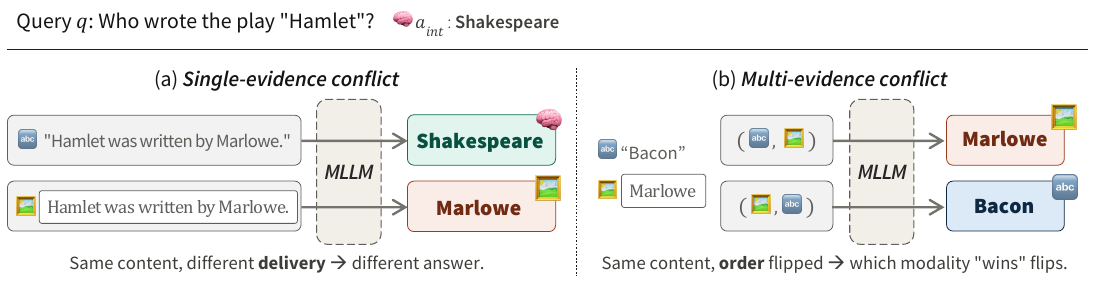}
    \caption{\textbf{The same evidence, delivered differently, can yield different answers.}
        \textit{(a) Single-evidence}: the model follows visual external evidence $e_I$(\img) but reverts to its memory $a_{\mathrm{int}}$(\brain) when the same content is delivered as text $e_T$(\txt).
        \textit{(b) Multi-evidence}: when $e_T$ and $e_I$ support different answers, 
        the order in which they appear can flip which modality the model follows.
        These examples illustrate a broader instability of MLLMs under knowledge conflict: the information a model prioritizes varies arbitrarily with input order, dataset, and model family.
        }
\label{fig:main_figure}
\end{figure*}

\section{Related Work}
\label{sec:related}

\subsection{Knowledge Conflict}
\label{sec:rw_kc}

Knowledge conflict has been studied as the tension between parametric memory and external evidence, formalized via entity substitution~\citep{longpre-etal-2021-entity} and later expanded along multiple conflict axes~\citep{xu-etal-2024-knowledge-conflicts, wang2024resolving}.
While these works characterize how models resolve such conflicts~\citep{jin-etal-2024-cutting, khandelwal-etal-2025-cocoa}, they generally assume a fixed modality, i.e., text.


The framework has been recently extended to multimodal settings where context is also presented as images \cite{zhangrobust, nguyen2025challenges, zhang2025modalities, hua2025vision}, but does not explicitly control for conflicts with the model's parametric memory.
Other studies test parametric or commonsense knowledge with counterfactual images, observing whether models follow the visual input or their internal beliefs \cite{liu-etal-2025-insight, ortu2025seeing}.
A related line examines inconsistencies between visual entities or attributes and their textual descriptions \cite{jia2026benchmarking}.
Across these studies, however, the evidence is predominantly natural images such as scenes or faces, rather than the knowledge-intensive sentences and paragraphs---presented either as text or as rendered images---that constitute the focus of our work.

\subsection{Modality Bias in MLLMs}
\label{sec:rw_modbias}

Recent work reports a \emph{``Blind Faith in Text''} tendency in MLLMs, where models favor text over visual evidence \cite{deng2025words}.
Counterfactual-image studies similarly show reliance on language priors over visual cues~\cite{lee-etal-2025-vlind}, and others report text-dominant behavior under controlled evidence conflicts \cite{zhang2025evaluating}.
A recent survey synthesizes these findings under the notion of modality collapse \cite{sim-etal-2025-vlms}.

We extend these efforts to knowledge-conflict scenarios, where consistent handling of input context is essential to the robust behavior of MLLMs.
We later show that the textual bias widely reported in the literature does not hold in our setting; the direction in fact often reverses, with MLLMs favoring images over text.
Furthermore, this trend grows more arbitrary as multiple pieces of evidence simultaneously contradict parametric knowledge, urging a reassessment of the current consensus.

\section{Problem Formulation}
\label{sec:problem_formulation}

In this work, we study \textit{modality robustness under knowledge conflict (KC)}, namely, whether MLLMs resolve knowledge conflicts consistently across different numbers and orderings of textual and visual inputs.
Let $p$ denote the model's internal parametric knowledge and $e$ the external evidence for query $q$. We specify two competing answer candidates: $a_{\mathrm{int}}$ supported by $p$ and $a_{\mathrm{ext}}$ supported by $e$, with $a_{\mathrm{int}} \neq a_{\mathrm{ext}}$.
To isolate modality, we hold the semantic content of the evidence fixed and present it either as text $e_T$ or as an image $e_I$.
Since this work focuses on knowledge-intensive scenarios, $e_I$ is defined as a rendered image of $e_T$.\footnote{The rendering procedure is provided in Appendix~\ref{app:rendering_method}.}
If the model produces different responses across $e_T$ and $e_I$, we regard this as a failure of modality robustness.

\subsection{Categories of Knowledge Conflict}
\label{subsec:conflict_settings}

Previous work typically focuses on either single- or two-evidence settings, often without carefully controlling for parametric knowledge. Our setup covers both configurations and explicitly accounts for parametric alignment.
Figure~\ref{fig:main_figure} illustrates the model's unstable responses in both settings, driven by inconsistent multimodal processing under KC.

\paragraph{Single-evidence conflict}
For each query $q$, we pair it with external evidence $e$ that contradicts $p$, instantiated in two modalities: textual $e_T$ and image $e_I$.
Both support the same answer $a_{\mathrm{ext}}$, conflicting with $a_{\mathrm{int}}$.
We compare two input conditions:
\[
x_{T} = (e_T, q) \quad \text{vs.} \quad x_{I} = (e_I, q).
\]
This measures how the modality of evidence affects the model's resolution of conflicts between parametric ($p$) and external ($e$) knowledge.

Specifically, we use KC datasets (\S\ref{subsec:experimental design}) with 5-tuples $(q, c^+, a^+, c_1^-, a_1^-)$: a query $q$, a factual context $c^+$ with answer $a^+$, and a counterfactual context $c_1^-$ with answer $a_1^-$.
For each $q$, we determine whether $a_\mathrm{int}$ matches $a^+$ or $a^-$, and align $(c^+, a^+)$ and $(c_1^-, a_1^-)$ with $(p, a_\mathrm{int})$ and $(e, a_\mathrm{ext})$.\footnote{Prior work on multimodal KC \cite{ortu2025seeing} either skipped this verification or assumed parametric memory is always correct, leaving their findings potentially confounded.}

To this end, we leverage a closed-book multiple-choice QA setup.
The answer choices are taken from the source KC dataset and augmented with a \textit{`none of the above'} option.
Inspired by \citet{wang2023selfconsistency}, we sample the model's output for $q$ five times and retain $q$ only when all five responses agree, taking the common answer as $a_\mathrm{int}$.\footnote{Unlike \citet{wang2023selfconsistency}, we require unanimous agreement across samples rather than majority voting, yielding a stricter estimate of internal knowledge.}
If $a_\mathrm{int}$ is \textit{`none of the above'}, we ask the model to provide an open-ended answer, which then replaces $a_\mathrm{int}$.
Subsequently, if $a_\mathrm{int} = a^+$, we align $(p, a_\mathrm{int})$ and $(e, a_\mathrm{ext})$ with $(c^+, a^+)$ and $(c_1^-, a_1^-)$, respectively; otherwise, the alignment is reversed.

As LLMs differ in their parametric knowledge, the number of retained instances ($|\mathcal{D}|$) after this process varies across models.
Model-wise filtering statistics are reported in Table \ref{tab:filter_stats_single} of Appendix \ref{app:dataset_filtering}.


\paragraph{Multi-evidence conflict}

We further consider a joint setting that provides both modalities to the model, in two possible orderings:
\[
x_{I \rightarrow T} = (e_I, e_T, q) 
\;\; \text{and} \;\;
x_{T \rightarrow I} = (e_T, e_I, q).
\]
We extend the setup to a three-way conflict by adding a second conflicting source, yielding 7-tuples $(q, c^+, a^+, c_1^-, a_1^-, c_2^-, a_2^-)$.
For simplicity, we focus on cases where $(p, a_\mathrm{int})$ matches the factual pair $(c^+, a^+)$.
We then have two possible assignments: (1) $(e_T, a_T) = (c_1^-, a_1^-)$ and $(e_I, a_I) = (c_2^-, a_2^-)$; (2) the reverse.
We average over both assignments and test whether the model favors one modality in this three-way conflict.
Again, model-wise final dataset sizes ($|\mathcal{D}|$) are reported in Table \ref{tab:filter_stats_multi} of Appendix \ref{app:dataset_filtering}.


\begin{table*}[!t]
    \centering
    \small
    \setlength{\tabcolsep}{4pt}
    \renewcommand{\arraystretch}{1.1}
\begin{tabular}{l rrr  rrr}
\toprule
\multirow{2}{*}{%
  \diagbox[width=9em,height=3em]
  {\textbf{Model}}{\textbf{Evidence}}%
}
& \multicolumn{3}{c}{\textbf{\textsc{ConflictQA}}}
& \multicolumn{3}{c}{\textbf{\textsc{NQ-Swap}}} \\
\cmidrule(lr){2-4}
\cmidrule(lr){5-7}

& \multicolumn{1}{c}{\textbf{Single}}
& \multicolumn{1}{c}{\textbf{Multi} $x_{I\rightarrow T}$}
& \multicolumn{1}{c}{\textbf{Multi} $x_{T\rightarrow I}$}
& \multicolumn{1}{c}{\textbf{Single}}
& \multicolumn{1}{c}{\textbf{Multi} $x_{I\rightarrow T}$}
& \multicolumn{1}{c}{\textbf{Multi} $x_{T\rightarrow I}$} \\
\midrule
Claude Sonnet 4.5
  & \posH  $42.86$ & \posVH $79.41$ & \posVH $84.62$
  & \posM  $22.63$ & \posL  $4.26$  & \posM  $19.38$ \\
GPT-5.4
  & \posM  $15.29$ & \posL  $1.80$  & \posVH $63.37$
  & \posL  $3.80$  & \posL  $4.38$  & \posL  $6.92$ \\
GPT-4o
  & \posL  $9.75$  & \posL  $4.40$  & \negM  $-19.05$
  & \negM  $-14.74$& \posL  $6.89$  & \posL  $4.80$ \\

\midrule
Qwen3-Omni
  & \posL  $1.64$  & \negH  $-38.24$ & \posVH $79.25$
  & \posM  $12.98$ & \negH  $-55.51$ & \posM  $10.94$ \\
Qwen2.5-Omni
  & \posL  $1.54$  & \negH  $-42.64$ & \posH  $43.77$
  & \posL  $6.74$  & \negL  $-5.64$  & \posH  $33.50$ \\
MiniCPM-o2.6
  & \posL  $1.09$  & \posL  $5.42$   & \posM  $26.36$
  & \posL  $1.86$  & \negM  $-26.18$ & \negH  $-42.27$ \\
OmniVinci
  & \negL  $-1.81$ & \negVH $-64.66$ & \posVH $86.58$
  & \negL  $-2.87$ & \posL  $0.94$   & \negL  $-0.77$ \\

\midrule
Qwen2.5-VL-32B
  & \posL  $2.32$  & \negM  $-19.37$ & \posVH $65.61$
  & \posL  $5.49$  & \negM  $-11.87$ & \posM  $16.96$ \\
Qwen2.5-VL-7B
  & \posL  $4.09$  & \negVH $-68.98$ & \posVH $73.82$
  & \posL  $3.80$  & \posH  $33.10$  & \posM  $29.51$ \\
InternVL3-38B
  & \posL  $0.46$  & \negM  $-21.32$ & \posVH $84.87$
  & \posL  $8.53$  & \negH  $-40.04$ & \posH  $42.92$ \\
InternVL3-8B
  & \negL  $-1.27$ & \negVH $-93.03$ & \posH  $58.77$
  & \posL  $3.44$  & \negH  $-47.25$ & \posM  $21.58$ \\
LLaVA-v1.6-34B
  & \negL  $-5.34$ & \negVH $-81.64$ & \negVH $-85.65$
  & \negL  $-9.75$ & \negH  $-31.38$ & \negH  $-43.28$ \\
LLaVA-vicuna-7B
  & \negL  $-9.17$ & \negVH $-95.45$ & \negVH $-94.86$
  & \negH  $-34.17$& \negVH $-78.76$ & \negVH $-75.67$ \\
\bottomrule
\end{tabular}
    \caption{
        Normalized image-over-text gap (\%) on \textsc{ConflictQA} and \textsc{NQ-Swap} across single- and multi-evidence settings; 
        $G^{\mathrm{single}}$ for single-evidence inputs,
        $G_o^{\mathrm{multi}}$ for multi-evidence inputs under each order $o \in \{I \rightarrow T, T \rightarrow I\}$.
        Cell shading scales with the magnitude of the change
        (\textcolor{blue!38}{$\blacksquare$}\,positive / \textcolor{red!38}{$\blacksquare$}\,negative;
        bins at $|x|\!\le\!10,\,10$--$30,\,30$--$60,\,>\!60$).
        }
    \label{tab:main_result}
\end{table*}

\subsection{Other Experimental Factors}
\label{subsec:experimental design}

We consider additional experimental factors that may affect modality-dependent evidence reliance. 

\paragraph{Datasets}
We evaluate on two complementary knowledge conflict datasets:
\begin{itemize}[itemsep=0pt, leftmargin=10pt]
    \item \textbf{\textsc{ConflictQA}} \cite{xie2024adaptive} constructs conflicts by eliciting the model's parametric memory and then generating counter-memory evidence supporting an alternative answer.
    Since the evidence is LLM-generated rather than produced by simple word substitution (as in \textsc{NQ-Swap}), the resulting passages read naturally.\footnote{We use the publicly available GPT-4-generated PopQA~\cite{mallen-etal-2023-trust} counterfactuals for $(c_1^-, a_1^-)$ and the ChatGPT-generated variant for $(c_2^-, a_2^-)$ or vice versa.}   
    \item \textbf{\textsc{NQ-Swap}} \cite{longpre-etal-2021-entity} is a variant of the Natural Questions dataset \cite{kwiatkowski-etal-2019-natural} where the answer entity is replaced with another entity in several substitution methods. 
    Since \textsc{NQ-Swap} is based on real QA passages, it provides a retrieval-like conflict setting.
\end{itemize}

\paragraph{Models}
We evaluate 13 MLLMs across proprietary API models and open-source model families:
\begin{itemize}[itemsep=0pt, leftmargin=10pt]
    \item \textbf{Proprietary models}: Claude Sonnet 4.5~\cite{4_5Sonnet}, GPT-5.4~\cite{openai2026gpt54}, and GPT-4o~\cite{hurst2024gpt}.
    \item \textbf{Open-source omni-modal LLMs (OLLMs)}: Qwen3-Omni-30B-A3B-Instruct~\cite{xu2025qwen3}, Qwen2.5-Omni-7B~\cite{xu2025qwen2}, MiniCPM-o2.6~\cite{openbmbminicpmo}, and OmniVinci~\cite{ye2025omnivinci}.
    \item \textbf{Open-source vision-language models (VLMs)}: Qwen2.5-VL-7B/-32B-Instruct~\cite{Qwen2.5-VL}, InternVL3-8B/-38B-Instruct~\cite{zhu2025internvl3}, LLaVA-v1.6-34b-hf, and LLaVA-v1.6-vicuna-7b-hf~\cite{liu2023visual}.
\end{itemize}
We test whether modality robustness under KC generalizes across model families and scales.

\paragraph{Metrics}

For $i\in\mathcal{D}$, we have distinct forms of prompts $\{x^{(i)}\}$, e.g., $x_T^{(i)}$ and $x_{I \rightarrow T}^{(i)}$.
We request an LLM to generate its open-ended response given a specific type of $x$ three times, and then evaluate them using normalized span matching.\footnote{We use normalized span matching to relax exact matching: after lowercasing, punctuation/whitespace cleanup, and alias handling, $\hat{a}_{x}^{(i)}$ counts as a match if either the normalized prediction contains $a_\mathrm{ext}^{(i)}$ or vice versa.}

Let $\hat{\mathbf{a}}_{x}^{(i)}$=$\{\hat{a}_{x,1}^{(i)},\hat{a}_{x,2}^{(i)},\hat{a}_{x,3}^{(i)}\}$ denote a set of three responses for instance $i$ under prompt $x$. 
We define the \textit{external-source following rate (EFR)} over the dataset $\mathcal{D}$ from \S\ref{subsec:conflict_settings}  as
\[
\mathrm{EFR}(x)\!
=\!
\frac{1}{|\mathcal{D}|}\!
\sum_{i\in\mathcal{D}}
\mathbb{I}\!\left[\hat{a}_{x,1}^{(i)} = \hat{a}_{x,2}^{(i)} = \hat{a}_{x,3}^{(i)} = a_{\mathrm{ext}}^{(i)}\right],
\]
where $\mathbb{I}\!\left[\hat{a}_{x,1}^{(i)} = \hat{a}_{x,2}^{(i)} = \hat{a}_{x,3}^{(i)} = a_{\mathrm{ext}}^{(i)}\right]$ becomes 1 only when all three responses match the external answer candidate $a^{(i)}_\mathrm{ext}$, and 0 otherwise.
By design, this definition is conservative: it credits a model only for consistently following the evidence, not for a single match that could arise by chance.

We quantify modality reliance using the normalized \textit{image-over-text gap}, denoted by $G$. 
Since models differ in their overall  \textit{EFR}, we normalize the image--text difference by their combined rate to enable stable cross-model comparison.
For single-evidence conflict, we compare two prompts that differ only in the modality of the same external evidence: $x_T$ and $x_I$. 
We thus compute $G$ as
\[
G^{\mathrm{single}}
=
\frac{\mathrm{EFR}(x_I)-\mathrm{EFR}(x_T)}
{\mathrm{EFR}(x_I)+\mathrm{EFR}(x_T)}
\times 100.
\]

In the multi-evidence setting, both $e_T$ and $e_I$ are presented jointly, supporting competing answers $a_T$ and $a_I$. For each ordering $o \in \{I \rightarrow T, T \rightarrow I\}$, 
we compute the modality-specific rates $\mathrm{EFR}_T(x_o)$ and $\mathrm{EFR}_I(x_o)$, and define the (order-aware) image-over-text gap as
\[
\qquad
G^{\mathrm{multi}}_o
=
\frac{\mathrm{EFR}_I(x_o)-\mathrm{EFR}_T(x_o)}
{\mathrm{EFR}_I(x_o)+\mathrm{EFR}_T(x_o)}
\times 100.
\]
The resulting gap is bounded within $[-100,100]$.
Positive values indicate that the model relies more on the image, with larger magnitudes corresponding to a wider image--text difference.
\section{MLLMs Lack Modality Robustness}
\label{sec:main_results}

Table \ref{tab:main_result} summarizes the normalized image-over-text gap $G$ across models, datasets, input orders, and conflict categories; the corresponding absolute percentage-point gaps are reported in Table~\ref{tab:app_efr_diff} in the Appendix.
The dominant pattern is \textit{instability}: modality is not merely a passive container of evidence, but a factor that influences how models resolve conflicts.

\paragraph{Single-evidence conflict: a weak but pervasive image preference.}

The single-evidence setting is the most direct test of presentation invariance: 
the external evidence has the same semantic content across conditions, and only its surface form changes.
Positive $G^{\mathrm{single}}$ therefore means that rendering the same evidence as an image makes the model more likely to follow it over its parametric knowledge.
This image advantage is most evident in proprietary models.
Claude Sonnet 4.5 shows the strongest positive gaps, with +42.86\% on \textsc{ConflictQA} and +22.63\% on \textsc{NQ-Swap}.
GPT-5.4 also remains positive on both datasets, with +15.29\% and +3.80\%, respectively. 
The pattern is not universal, however: GPT-4o reverses from a positive gap on \textsc{ConflictQA} (+9.75\%) to a negative gap on \textsc{NQ-Swap} (-14.74\%), while the LLaVA models show consistently negative gaps, indicating stronger reliance on text evidence.
These results show that even the simplest presentation change can alter evidence following.

\paragraph{Multi-evidence conflict: modality order matters.}
Table \ref{tab:main_result} shows that when $e_I$ precedes $e_T$ ($x_{I \rightarrow T}$), most open-source models show negative gains, 
indicating stronger reliance on the text-supported answers.
However, when the order is reversed ($x_{T\rightarrow I}$), the same models often shift toward image-supported answers.
This order sensitivity is reflected in both the per-model \textit{Flip Ratio} and the aggregate number of sign flips, as shown in Table \ref{tab:app_flip_ratio} in the Appendix.
The \textit{Flip Ratio} further shows instance-level preference changes when the input order is reversed.
At the aggregate level, 8 out of 13 models on \textsc{ConflictQA} and 6 out of 13 models on \textsc{NQ-Swap} change the sign of $G$ across the two input orders.
These reversals show that modality reliance is not fixed; it can be reshaped by the order in which conflicting sources enter the context.

\paragraph{$G$ is also influenced by the dataset.}
As Table~\ref{tab:main_result} shows in the multi-evidence settings, \textsc{ConflictQA} elicits markedly larger absolute gaps than \textsc{NQ-Swap};
OmniVinci, for example, exceeds 86\% on \textsc{ConflictQA} but is nearly neutral on \textsc{NQ-Swap}. 
We hypothesize that modality-dependent reliance grows with the semantic divergence between conflicting sources. 
\textsc{NQ-Swap} perturbs only the answer entity and otherwise leaves the passage intact, so the two evidence sources remain largely overlapping; \textsc{ConflictQA}, by contrast, generates entirely separate counter-evidence, producing a much broader semantic gap.
Figure~\ref{fig:app_dataset-divergence} in Appendix~\ref{app:dataset_filtering} supports this characterization.

\begin{figure}[t]
\centering
\includegraphics[width=\columnwidth]{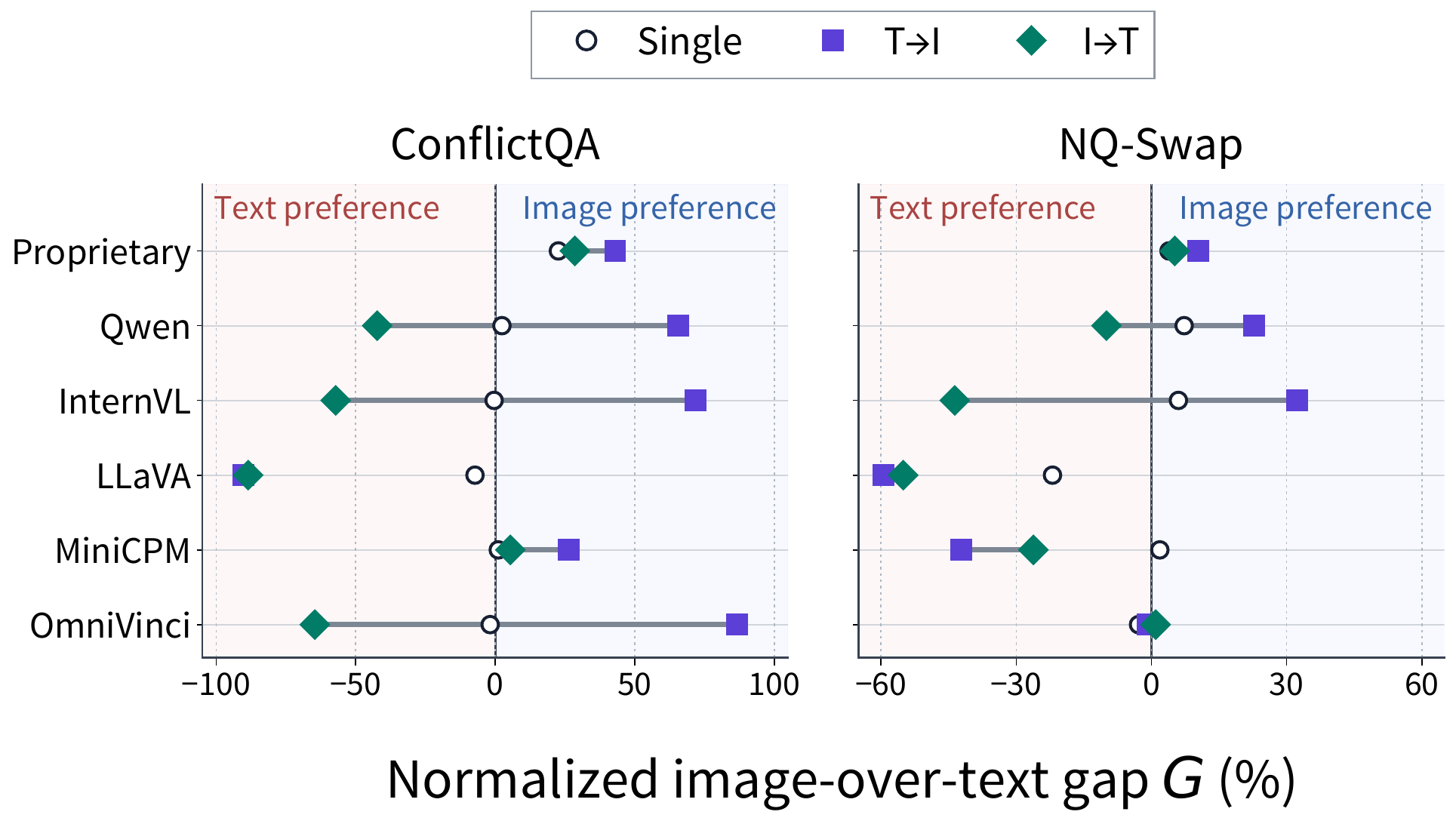}
    \caption{
    Group-averaged modality preference on \textsc{ConflictQA} and \textsc{NQ-Swap}.
    Positive and negative values indicate image and text preference, respectively. 
    }
\label{fig:main_visualization}
\end{figure}

\paragraph{Variation across model families.}
Figure~\ref{fig:main_visualization} visualizes $G$ from Table~\ref{tab:main_result} 
by grouping open-source models by family and proprietary models into a separate category.
The proprietary group maintains the most consistently positive average $G$, indicating a relatively stable image preference. 
In contrast, LLaVA models remain strongly text-dominant across all conditions and datasets. 
Qwen and InternVL families exhibit pronounced order sensitivity: their average $G$ becomes strongly positive under $x_{T\rightarrow I}$ but negative under $x_{I\rightarrow T}$, especially on \textsc{ConflictQA}. 
Together, these patterns indicate that modality-dependent reliance can be influenced by model characteristics and evidence conditions, rather than reflecting a universal property of MLLMs.

\section{When Modality Instability Matters}



The instability documented in \S\ref{sec:main_results} is not merely a benchmark artifact.
Deployed MLLM pipelines rarely control how evidence arrives. 
Even when the underlying evidence remains identical, models may resolve conflicts differently due to upstream choices in evidence representation.

We examine two settings where this failure mode is consequential: question-answering performance in multimodal RAG (\S\ref{subsec:rag}), where format changes what the model attends to, and safety vulnerability to image-rendered adversarial instructions (\S\ref{subsec:safety}).

\subsection{Impact of Gold-Passage Modality in RAG}
\label{subsec:rag}


To probe whether modality instability matters in practice, we begin with retrieval-augmented QA. 
As a thought experiment, imagine that an ideal paragraph directly answering the query is given in advance: how do MLLMs treat this gold-standard passage when only its modality changes?
We fix the query, the gold passage's content, and the distractors (all provided as text for simplicity), and vary only the modality of the gold passage.

We sample 170 query--passage instances from MS MARCO v2.1~\citep{bajaj2016ms} and report answer quality with ROUGE-L. 
MLLMs are evaluated under four cases: a \textit{gold-only} setting, in which the gold passage is the only input, and three \textit{multi-passage} settings, in which it appears among nine distractors at the \textit{First}, \textit{Middle}, or \textit{Last} position.

\begin{figure}[t]
\centering
\includegraphics[width=\columnwidth]{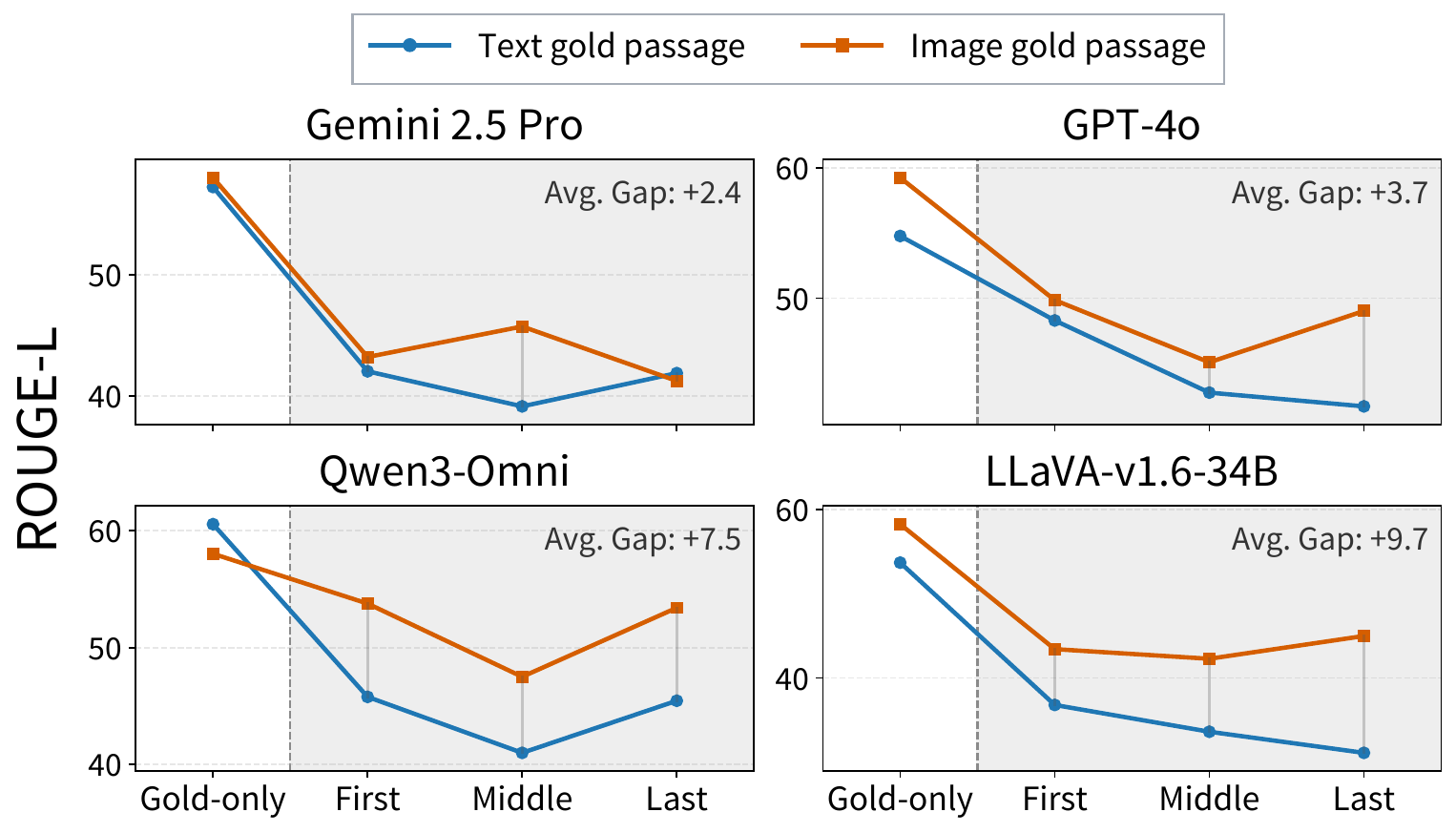}
    \caption{
    RAG performance on MS MARCO v2.1 with text vs. image gold passages.
    \textit{Gold-only}: gold alone; \textit{First/Middle/Last}: gold position among nine distractors. 
    \textit{Avg. Gap}: the mean image--text ROUGE-L difference.}
\label{fig:rag}
\end{figure}


Figure~\ref{fig:rag} demonstrates that the modality of the gold-standard passage influences RAG performance, with image rendering mitigating the well-known lost-in-the-middle phenomenon \cite{liu-etal-2024-lost}.
For instance, Gemini 2.5 Pro gains +6.59 ROUGE-L when the image-rendered gold passage occupies the middle position. 
This suggests that rendering relevant context as an image may serve as a new strategy to direct model attention toward salient passages embedded among distractors.

\subsection{Vulnerability to Image-Rendered Attacks}
\label{subsec:safety}


As a second case study, we ask whether modality instability also affects MLLMs' safety.
Our experiments build on MM-SafetyBench \citep{liu2024mm}, which benchmarks robustness against attacks that hide harmful content in query-relevant images. 
Its four conditions---Text-Only, Stable Diffusion (SD), Typography (Typo), and SD+Typo---serve as our baselines and embed only the key phrase in image form.\footnote{SD generates images from prompts, while Typo overlays the key phrase on a white background; see Appendix~\ref{app:safety}.} 
We extend this setup with an \textit{image-rendered instruction} condition, in which the full instruction is moved into the visual channel, leaving the request semantically unchanged, and report Attack Success Rate (ASR), the proportion of prompts for which the model produces harmful content instead of refusing.

\begin{figure}[t]
\centering
\includegraphics[width=\columnwidth]{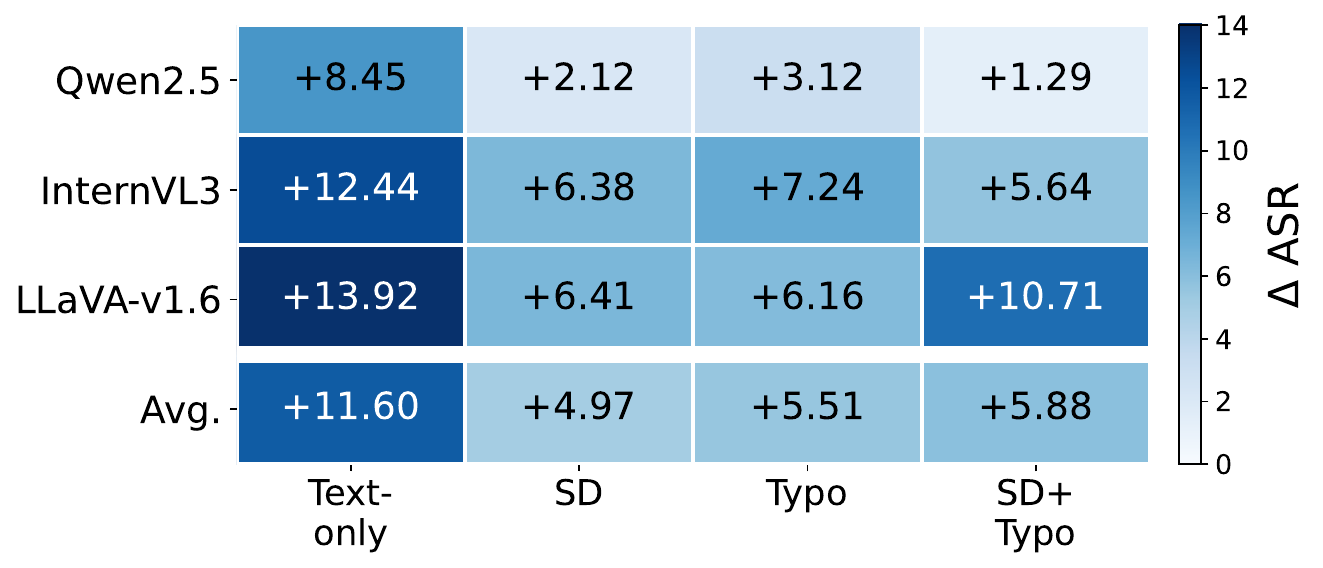}
    \caption{Mean $\Delta$ASR over 13 MM-SafetyBench scenarios when the harmful instruction is rendered as an image.
    Cells report the points relative to each baseline (Text-Only, SD, Typo, SD+Typo); darker shading indicates larger increases.
    Avg. averages over the three models; per-scenario values in Appendix~\ref{app:safety}.}
\label{fig:ASR}
\end{figure}


Figure~\ref{fig:ASR} summarizes the increase in ASR from rendering harmful instructions as images across models and baseline conditions, averaged over all 13 scenarios.
Over 156 evaluation settings in total, this intervention increases ASR by 6.99\% on average, with positive $\Delta$ASR in 126/156 (80.8\%) comparisons.
Table~\ref{tab:safety_full} in the Appendix reports the corresponding per-scenario results.
The average increase is consistently positive across all model $\times$ baseline cells, although its magnitude varies across models (Qwen +3.74\%, InternVL +7.93\%, LLaVA +9.30\%). 
It is largest against the Text-only baseline (+11.60\%) and smallest against SD (+4.97\%). 
Together, these results suggest that moving the whole instruction into the visual channel introduces an additional safety vulnerability beyond existing attacks that embed only parts of the instruction visually.





\section{Mitigating Modality Instability}
\label{sec:mitigation}

Having established the modality instability of MLLMs and its practical consequences, we next explore whether this behavior can be mitigated.
We consider interventions at three levels: a lightweight intervention through explicit prompting, post-training through supervised fine-tuning and direct preference optimization, and inference-time control through representation steering.


\subsection{Prompting is Not Sufficient}

A natural first step is to instruct the model directly to use both modalities in balance.
We therefore prepend a system-level prompt asking the model to jointly consider textual and visual evidence whenever multiple sources are provided. 
Under the hypothesis that underspecified instructions drive the imbalance, this explicit multimodal prompting should close the gap.

Contrary to expectation, as shown in Table~\ref{tab:prompting}, prompting does not provide reliable control over modality reliance.
While it partially reduces the imbalance for Qwen2.5-VL-7B, it instead amplifies the bias for GPT-5.4. 
We observe similar instability when the prompt explicitly asks the model to focus on a particular modality. 
These outcomes suggest that modality-dependent reliance cannot be reliably corrected through surface-level prompting alone, motivating the post-training and inference-time interventions explored next.

\begin{table}[t]
    \centering
    \footnotesize
    \begin{tabular}{l c >{\columncolor{gray!10}}c c > {\columncolor{gray!10}}c}
    \toprule
        & \multicolumn{2}{c}{$\textrm{Multi}_{I\rightarrow T}$ }
        & \multicolumn{2}{c}{$\textrm{Multi}_{T\rightarrow I}$} \\
    \cmidrule(lr){2-3} \cmidrule(lr){4-5}
        \textbf{Model} 
        & \textbf{Orig.} & \textbf{+Prompt.}
        & \textbf{Orig.} & \textbf{+Prompt.} \\
    \midrule
        GPT-5.4 & 1.80 & 71.26 & 63.37 & 80.49 \\
        Qwen2.5-VL & -68.97 & -17.54 & 73.82 & 45.69 \\
    \bottomrule
    \end{tabular}
    \caption{
    Image-over-text gap ($G^{\mathrm{multi}}_o$, \%) before and after explicit prompting on the \textsc{ConflictQA} subset in the multi-evidence setting. 
    Values closer to zero indicate more balanced reliance on textual and visual evidence.
    }
    \label{tab:prompting}
\end{table}


\subsection{Conflict-Aware Fine-Tuning}
\label{subsec:sft}

We here examine whether this modality-dependent reliance can be recalibrated through post-training. 
Unlike prompting, which intervenes only through instructions at inference time, fine-tuning directly adjusts the model's behavior under conflicting multimodal evidence.

To test whether reliance on misleading visual evidence can be recalibrated in a reproducible open-source setting, we select Qwen2.5-VL-7B-Instruct, which exhibits strong image-following behavior.
We then apply conflict-aware LoRA fine-tuning using training data constructed from \textsc{NQ-Swap} and MMM-Fact \citep{xu2025mmm}.\footnote{For \textsc{NQ-Swap}, we train the model to produce the ground-truth answer despite a counterfactual image-rendered passage. For MMM-Fact, we use non-supporting image-description pairs, where the image and the textual description disagree, to encourage cross-modal comparison. See Appendix~\ref{app:app_sft}.}
This tuning objective reduces the model’s tendency to follow misleading image evidence while preserving its use of factual evidence during conflict resolution.

As shown in Table~\ref{tab:sft}, conflict-aware SFT consistently reduces the magnitude of the modality gap across all evaluated settings. 
The effect is particularly pronounced in the multi-evidence $x_{T\rightarrow I}$ condition, where the image-dominant gap decreases from 73.82\% to 52.04\%.
The opposite order shows a modest reduction in the text-dominant gap. 
The intervention also reduces modality imbalance in the single-evidence setting for both misleading (-2.29\%) and reliable (-2.37\%) evidence.

Importantly, moving $G$ toward zero does not by itself guarantee desirable evidence use, since balanced modality reliance could in principle arise from uniformly increasing or decreasing evidence following. 
We therefore additionally examine the corresponding raw $\mathrm{EFR}_I$ in Appendix~\ref{app:app_sft}. 
Table~\ref{tab:app_sft} shows that SFT reduces following of misleading image evidence while preserving, and slightly increasing, following of reliable image evidence. 
These findings provide evidence that conflict-aware SFT can improve modality balance without merely suppressing visual evidence or shifting preference toward the opposite modality.


\begin{table}[t]
    \centering
    \small
    \setlength{\tabcolsep}{5pt}
    \renewcommand{\arraystretch}{1.05}
        \begin{tabular}{lcccc}
        \toprule
            \textbf{Evidence}
            & \multicolumn{3}{c}{\textbf{False}}
            & \textbf{True} \\
            \cmidrule(lr){2-4}
            \cmidrule(lr){5-5}
        
            \textbf{Method}
            & $\textrm{Multi}_{I\rightarrow T}$
            & $\textrm{Multi}_{T\rightarrow I}$
            & $\textrm{Single}$
            & $\textrm{Single}$ \\
        \midrule
        
        Original & -68.97 & 73.82 & 2.54 & 3.45 \\
        \rowcolor{gray!10}
        + SFT & -67.07 & 52.04 & 0.25 & 1.08 \\
        
        \bottomrule
        \end{tabular}
        
    \caption{
    Effect of conflict-aware fine-tuning on Qwen2.5-VL-7B on \textsc{ConflictQA}.
    False and True denote misleading and reliable evidence, respectively.}
    \label{tab:sft}
\end{table}

\subsection{Additional Mitigation Baselines}

To broaden our mitigation analysis beyond prompting and conflict-aware SFT, we additionally evaluate two approaches operating at different stages: 
Direct Preference Optimization (DPO) \cite{rafailov2023direct} as a post-training method and representation-level steering as an inference-time intervention. 
Detailed implementation settings and the corresponding raw image-following rates are provided in Appendix~\ref{app:app_dpo} and \ref{app:app_steering}.

\paragraph{Preference-based post-training.} 
We construct \textsc{NQ-Swap} preference pairs that favor the original factual answer over the answer supported by the substituted misleading image, and use them to train Qwen2.5-VL-7B-Instruct with DPO.
Table~\ref{tab:additional_mitigation} reports that DPO slightly reduces the image-dominant gap under $x_{T\rightarrow I}$, but substantially increases the text-dominant gap under $x_{I\rightarrow T}$.
Thus, DPO can shift reliance away from misleading image evidence in some cases, but does not improve modality balance across different input orders.


\paragraph{Representation-level steering.}
We further evaluate an inference-time steering baseline following prior work~\cite{zhang2025evaluating} on the same Qwen2.5-VL-7B model. 
Using a calibration subset of \textsc{ConflictQA}, we derive separate text-following steering directions for each input order and apply them during decoding on a held-out evaluation subset.
As the steering strength increases, the image-dominant gap under $x_{T\rightarrow I}$ decreases from 77.77\% to 60.00\%. 
In contrast, the text-dominant gap under $x_{I\rightarrow T}$ grows. 
At stronger steering scales, modality preference shifts toward text under both input orders.

Overall, both methods exhibit a directional trade-off: 
reducing image-dominant behavior can coincide with stronger text-dominant behavior.
This contrasts with conflict-aware SFT, which moves every $G$ closer to zero shown in Table~\ref{tab:sft}. 
These observations highlight an important distinction between controlling modality preference and improving modality robustness; 
effective mitigation should reduce imbalance in both directions rather than simply shifting reliance toward a fixed modality.

\begin{table}[t]
\centering
\small
\setlength{\tabcolsep}{10pt}
    \begin{tabular}{lcc}
    \toprule
    \textbf{Method}
    & $\textrm{Multi}_{I\rightarrow T}$
    & $\textrm{Multi}_{T\rightarrow I}$ \\
    \midrule
    
    \multicolumn{3}{l}{\textbf{Preference-based post-training}} \\
    Original & -68.89 & 68.47 \\
    \rowcolor{gray!10}
    + DPO & -88.08 & 65.57 \\
    
    \midrule
    \multicolumn{3}{l}{\textbf{Representation-level steering}} \\
    Original (Scale 0) & -38.64 & 77.77 \\
    \rowcolor{gray!10}
    Scale 1 & -45.05 & 80.64 \\
    \rowcolor{gray!10}
    Scale 3 & -70.21 & 64.95 \\
    \rowcolor{gray!10}
    Scale 5 & -86.96 & 60.00 \\
    
    \bottomrule
    \end{tabular}
    
    \caption{
    Additional mitigation interventions on multi-evidence conflicts.
    Positive and negative $G_o^{\mathrm{multi}}$ indicate image- and text-dominant reliance, respectively.
    }
    \label{tab:additional_mitigation}
\end{table}
\section{Further Analysis: Input-Side Sensitivity}


Given that input processing is the stage where modality-specific handling is most pronounced, we ask whether input-side interventions---altering how visual evidence is preprocessed and rendered---can change a model's modality reliance.
We examine this with frozen model weights, intervening solely on the visual preprocessing pipeline and the rendered form of semantically identical evidence.

\subsection{Influence of Visual Preprocessing}
\label{subsec:preprocessing}

The results in \S\ref{sec:main_results} show that modality reliance varies markedly across model families. 
To probe one possible source of this variation, we focus on Qwen2.5-VL and LLaVA-v1.6-7B, two families with opposing tendencies: Qwen leans toward image evidence, LLaVA leans toward text. 
Using this contrast, we examine whether preprocessing-level interventions can shift the model away from modality-dependent instability.
We apply bidirectional interventions between the two models. 
Specifically, Qwen2.5-VL receives LLaVA-style 336$\times$336 global/local view construction, 
whereas LLaVA-v1.6 receives images pre-resized by Qwen's smart-resize rule.
All remaining model-specific processing is unchanged; see Appendix~\ref{app:visual_preprocessing_intervention} for details.

\begin{table}[t]
\centering
\small
\setlength{\tabcolsep}{5pt}
    \begin{tabular}{llc>{\columncolor{gray!10}}c}
    \toprule
    \textbf{Base Model}
    & \textbf{Preprocessing}
    & \textbf{Orig.}
    & \textbf{After Prep.} \\
    \midrule
    Qwen2.5-VL & LLaVA-style & 9.71 & -37.23 \\
    LLaVA-v1.6 & Qwen-style & -34.92 & -30.91 \\
    \bottomrule
    \end{tabular}
    \caption{
    Effect of visual preprocessing on $G^{\mathrm{single}}$ for \textsc{NQ-Swap}. 
    \textit{Orig.} is the original model result; \textit{After Prep.} is the result after each preprocessing intervention.
    }
    \label{tab:visual_preprocessing}
\end{table}

Table~\ref{tab:visual_preprocessing} reveals a clear but asymmetric effect across intervention directions.
Under LLaVA-style preprocessing, Qwen's $G^{\mathrm{single}}$ moves from 9.71\% to -37.23\%, not merely flipping sign but landing within three percentage points of LLaVA's native value (-34.92\%). 
Input-side processing alone is thus sufficient to reproduce LLaVA's modality profile in a model with different weights and training. 
The reverse intervention produces only a modest shift, from -34.92\% to -30.91\%.

The key implication of this analysis is that preprocessing is one factor in modality reliance but not the sole mechanism; our setup also cannot isolate which downstream components interact with it.
A more comprehensive test would require training matched models that differ only in preprocessing, which we leave to future work.

\subsection{Influence of Image Presentation}
\label{subsec:image_color}

A natural concern is whether our findings depend on arbitrary choices in how textual evidence is rendered as images. 
To address this, we perturb the default rendering at two levels of increasing visual salience and measure how the model's source-following behavior shifts. 

On the \textsc{NQ-Swap} subset, we change only the visual appearance of the rendered evidence, leaving the query and evidence content untouched.
The first level applies \textit{global} changes to the entire image, either by rendering all text in red or by setting a yellow background.
The second level applies the same changes only to the \textit{targeted} answer-bearing span; this serves as an upper-bound salience diagnostic rather than a deployable setting, since it presupposes knowledge of the answer.

\begin{figure}[t]
\centering
\includegraphics[width=\columnwidth]{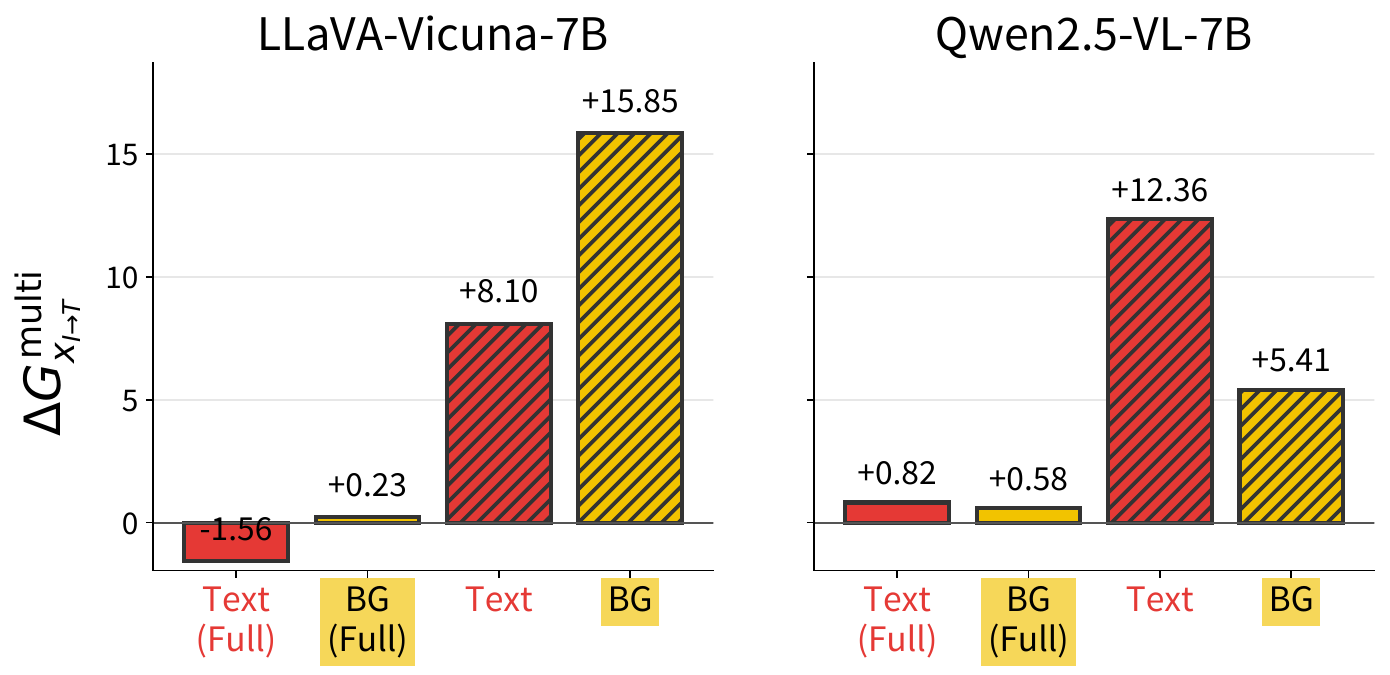}
    \caption{
    Effect of visualization variants on $G_{x_{I\!\rightarrow\!T}}^{\mathrm{multi}}$ on \textsc{NQ-Swap}, relative to the default value. 
    \textit{\textcolor{red}{Text}} and \textit{\colorbox{yellow}{BG}} highlight the answer-bearing span (red text and yellow background, respectively). 
    \textit{(Full)} applies the same changes to the entire evidence.
    }
\label{fig:image_color}
\end{figure}

Figure~\ref{fig:image_color} shows that modest global visual changes produce only small shifts relative to the default rendering, whereas answer-span highlighting produces much larger shifts.
These findings both strengthen and qualify our main finding:
the modality instability documented in \S\ref{sec:main_results} is not an artifact of font, color, or background choice, since reasonable variations along those dimensions leave the gain magnitudes largely intact. 
Yet the answer-span result indicates that image evidence carries an additional axis of variability that text evidence does not. 
This aligns with the effectiveness of image-rendered attacks observed in \S\ref{subsec:safety}, suggesting that the influence of image evidence is shaped not only by its content but also by its visual form.

\section{Conclusion}

We study whether MLLMs resolve knowledge conflicts consistently across modalities and find that they do not: the same conflict yields different answers depending on how the evidence is delivered, and the direction of preference shifts with dataset, input order, and evidence composition.
The fixed text or image preference reported in prior work does not survive once evidence is allowed to vary along these axes, and no single factor we examine reliably predicts which modality a model will favor.
\textit{Modality preference} is thus better understood as an artifact of the evaluation setup than as a fixed property of the model.

This instability has practical consequences, both beneficial (e.g., highlighting salient passages in RAG) and harmful (e.g., more potent adversarial attacks); the simple remedies we explored were largely ineffective, with conflict-aware fine-tuning offering only partial mitigation.
The origin of this instability remains an open question.
Our preliminary findings identify the visual input pipeline as a tractable starting point, but a complete account will require examining multiple stages of model training, which we leave to future work.

\section*{Limitations}
First, we consider only text-modality evidence and its rendered-image form, and do not examine naturally occurring visual evidence (e.g., figures, charts, photographs) 
or other modalities such as audio, where modality dynamics may differ.
Extending our framework to these broader modalities is a direction for future work. 
Second, while we evaluate 13 MLLMs across proprietary and open-source families, our coverage is not exhaustive.
Third, while we identify several contributing factors, a more thorough mechanistic account of why modality reliance shifts in the ways we document remains to be developed.

\section*{Ethical Statement}
Our safety experiments are intended to diagnose modality-dependent weaknesses in MLLM refusal behavior, not to provide new attack recipes. 
We report aggregate ASR results and do not include concrete harmful instructions or unsafe generations. 
Any released artifacts will exclude harmful rendered prompts and will be limited to evaluation metadata and non-sensitive analysis code.

\section*{Acknowledgments}
This work was supported by Institute of Information \& communications Technology Planning \& Evaluation (IITP) grant funded by the Korea government(MSIT) (No.RS-2020-II201373, Artificial Intelligence Graduate School Program(Hanyang University)).
This work was supported by Institute of Information \& communications Technology Planning \& Evaluation (IITP) under the artificial intelligence semiconductor support program to nurture the best talents (IITP-2026-RS-2023-00253914) grant funded by the Korea government(MSIT).
This work was supported by the National Research Foundation of Korea(NRF) grant funded by the Korea government(MSIT) (RS-2025-00558151).

\bibliography{anthology_part1, anthology_part2, custom}

\appendix
\newpage
\section*{Appendix}


\section{Evaluation Data and Input Construction}
\subsection{Text-to-Image Rendering}
\label{app:rendering_method}
We render textual evidence as image using the Python Pillow library. 
Unless otherwise specified, the following rendering template is adopted: black text on a white background, with a font family DejaVuSans, a font size of 24, a maximum text width of 800 pixels, and a line-spacing ratio of 1.25. 
The rendered text is center-aligned within the image.
Line breaks are determined by rendered pixel width: words are wrapped to the next line when they exceed the maximum width, and overlong words are split at the character level. 
The final image size is then set adaptively from the wrapped text bounding box, with an additional 8-pixel padding to prevent clipping.

Preliminary checks showed that changes in font size(24-32) and text width(600-1,000) had little effect on model behavior. 
We therefore fix the rendering template in the main experiments and focus on the effect of delivery modality.

\subsection{Dataset Preprocessing and Instance Selection}
\label{app:dataset_filtering}
We use two knowledge-conflict datasets, \textsc{ConflictQA} and \textsc{NQ-Swap}, and apply lightweight preprocessing before constructing our conflict instances.

\paragraph{\textsc{ConflictQA}.}
We use the GPT-4-generated version of \textsc{ConflictQA}, which is constructed from PopQA and contains 9,544 instances. 
Each instance provides an original answer--context pair and a counter-answer--context pair. 
We exclude instances whose evidence context contains non-informative responses such as ``don't know'' or ``I cannot answer'', since such contexts do not provide usable external evidence for a conflicting answer.

\paragraph{\textsc{NQ-Swap}.}
\textsc{NQ-Swap} comprises 4,746 Natural Questions variants in which the answer entity in each original context is replaced with another entity to create conflicting evidence.
We exclude contexts containing excessive markup, particularly repeated \texttt{<Table>} tags, because they disrupt the linear reading order and produce cluttered or structurally ambiguous image renderings, making it difficult to maintain semantic alignment between the text and rendered-image conditions.

\paragraph{Model-specific filtering.}
After preprocessing, we identify the memory-supported candidate $a_{\mathrm{int}}$ separately for each model using the closed-book prediction procedure described in \S\ref{sec:problem_formulation} and Figure~\ref{fig:prompt_internal}. 
We retain only instances with consistent predictions across five runs and a valid opposing answer--context pair that can serve as external evidence. 
Therefore, the final number of usable single-evidence conflict pairs differs by model. 
Table~\ref{tab:filter_stats_single} and Table~\ref{tab:filter_stats_multi} report the retained instance counts for each model and dataset.

\paragraph{Multi-evidence input extension.}
For multi-evidence experiments, we require two distinct external evidence sources for the same query. 
For \textsc{ConflictQA}, we use the ChatGPT-generated variant in addition to the default GPT-4-generated data, and use a differently tagged context for the same query as the second external input. 
For \textsc{NQ-Swap}, when multiple substituted contexts are available for the same original context, we use another substitution as the second external evidence source. 
Table~\ref{tab:filter_stats_multi} reports the number of unique multi-evidence pairs constructed in this way.
For each pair, we evaluate both modality assignments, swapping which context is presented as text and which is rendered as an image.
Thus, the number of evaluated cross-modal inputs is twice the number of unique pairs in Table~\ref{tab:filter_stats_multi} for each presentation order.

\begin{table}[t]
    \centering
    \small
    \begin{tabular}{lrr}
    \toprule
        \textbf{Model} & \textbf{\textsc{ConflictQA}} & \textbf{\textsc{NQ-Swap}} \\
    \midrule
        Claude Sonnet 4.5 & 1,198 & 3,205 \\
        GPT-5.4 & 2,695 & 3,270 \\
        GPT-4o & 4,049 & 3,310\\
        Qwen3-Omni & 1,670 & 3,047 \\
        Qwen2.5-Omni & 694 & 2,532\\
        MiniCPM-o2.6 & 1,170 & 2,863\\
        OmniVinci & 626 & 2,780\\
        Qwen2.5-VL-32B & 1,019 & 2,904 \\
        Qwen2.5-VL-7B & 718 & 2,169\\
        InternVL3-38B & 740 & 2,859 \\
        InternVL3-8B & 676 & 2,723\\
        LLaVA-v1.6-34B & 589 & 3,050 \\
        LLaVA-v1.6-7B & 907 & 2,671 \\
    \bottomrule
    \end{tabular}
    \caption{
    Model-wise filtering statistics for constructing single-evidence conflict pairs in \textsc{ConflictQA} and \textsc{NQ-Swap} (original: 9,544 and 4,746 respectively). 
    Each value denotes the number of retained instances after five closed-book predictions and validity filtering.}
    \label{tab:filter_stats_single}
\end{table}

\begin{table}[t]
    \centering
    \small
    \begin{tabular}{lrr}
    \toprule
        \textbf{Model} & \textbf{\textsc{ConflictQA}} & \textbf{\textsc{NQ-Swap}} \\
    \midrule
        Claude Sonnet 4.5 & 442 & 825 \\
        GPT-5.4 & 972 & 792 \\
        GPT-4o & 885 & 727 \\
        Qwen3-Omni & 313 & 743 \\
        Qwen2.5-Omni & 181 & 632 \\
        MiniCPM-o2.6 & 344 & 732 \\
        OmniVinci & 199 & 716 \\
        Qwen2.5-VL-32B & 253 & 721  \\
        Qwen2.5-VL-7B & 135 & 461 \\
        InternVL3-38B & 171 & 721 \\
        InternVL3-8B & 184 & 703 \\
        LLaVA-v1.6-34B & 226 & 802 \\
        LLaVA-v1.6-7B & 335 & 686 \\
    \bottomrule
    \end{tabular}
    \caption{Model-wise filtering statistics for constructing multi-evidence conflict pairs in \textsc{ConflictQA} and \textsc{NQ-Swap}.
    Considering the cross-modal source assignments, the number of evaluated inputs is the twice of these unique pairs.}
    \label{tab:filter_stats_multi}
\end{table}

\paragraph{Dataset-Level Evidence Divergence}
\label{app:dataset-divergence}

To examine the dataset-dependent modality gap, we quantify the divergence between the paired evidence passages within each dataset. 
For \textsc{ConflictQA}, we compare the separately generated \texttt{parametric\_memory} and \texttt{counter\_memory} passages, which support the parametric and counter answers, respectively. 
For \textsc{NQ-Swap}, we compare \texttt{org\_context} with \texttt{sub\_context}, where the latter is constructed by replacing the answer entity in the original passage.

\begin{figure}[t]
    \centering
    \includegraphics[width=\columnwidth]{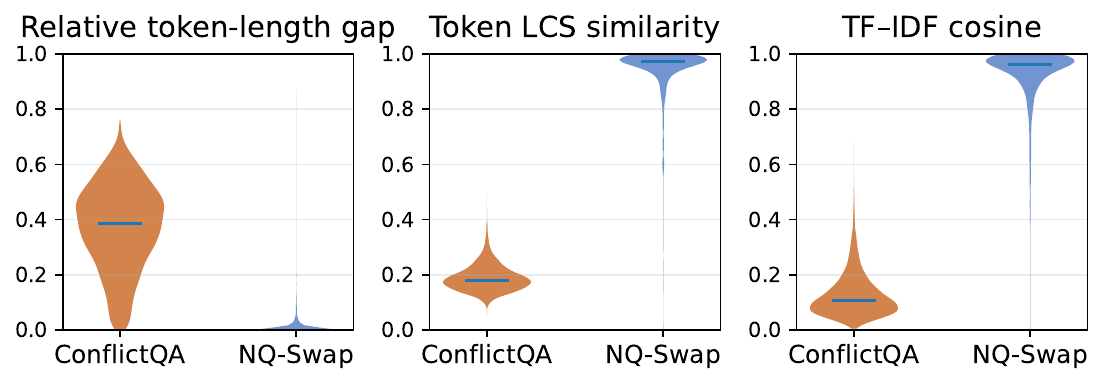}
    \caption{Distributions of relative token-length gap, token LCS similarity, and TF--IDF cosine similarity between paired evidence passages in \textsc{ConflictQA} and \textsc{NQ-Swap}.
    Horizontal bars denote means.}
    \label{fig:app_dataset-divergence}
\end{figure}

As shown in Figure~\ref{fig:app_dataset-divergence}, \textsc{ConflictQA} exhibits a much larger relative token-length gap than \textsc{NQ-Swap} (0.372 vs.\ 0.010), while showing substantially lower token LCS similarity (0.188 vs.\ 0.942) and TF--IDF cosine similarity (0.126 vs.\ 0.922). 
Moreover, after masking the paired answer strings (\texttt{memory\_short\_answer}/\texttt{counter\_short\_answer} for \textsc{ConflictQA} and \texttt{org\_answer}/\texttt{sub\_answer} for \textsc{NQ-Swap}), 97.9\% of \textsc{NQ-Swap} passage pairs become identical, compared with 0.0\% of \textsc{ConflictQA} pairs. 
The overlap difference also remains after matching examples by passage length.
These observations indicate that \textsc{NQ-Swap} primarily applies a localized answer substitution, whereas \textsc{ConflictQA} constructs globally distinct counter-evidence.

\section{Extended Main Results}
\label{app:ext}
\subsection{Order Sensitivity in Multi-evidence Conflicts}
Beyond aggregate modality preference in Table~\ref{tab:main_result}, we further examine whether individual predictions are sensitive to evidence ordering. 
Table~\ref{tab:app_flip_ratio} reports the flip ratio when reversing the order of multimodal evidence. 
A high flip ratio indicates that the model does not consistently prioritize one evidence source, but changes its preference depending on presentation order.

\begin{table}[!t]
    \centering
    \small
    \setlength{\tabcolsep}{4pt}
    \renewcommand{\arraystretch}{1.1}
    \begin{center}
\small
\setlength{\tabcolsep}{6pt}

\begin{tabular}{lcc}
    \toprule
    \textbf{Model}
    & \textbf{\textsc{ConflictQA}}
    & \textbf{\textsc{NQ-Swap}} \\
    \midrule

    Claude Sonnet 4.5 & 4.35  & 21.47 \\
    GPT-5.4           & 16.10 & 17.59 \\
    GPT-4o            & 19.14 & 14.48 \\

    \midrule
    Qwen3-Omni        & 48.57 & 26.95 \\
    Qwen2.5-Omni      & 38.77 & 33.01 \\
    MiniCPM-o2.6      & 23.70 & 14.68 \\
    OmniVinci         & 66.81 & 34.07 \\

    \midrule
    Qwen2.5-VL-32B    & 33.33 & 27.78 \\
    Qwen2.5-VL-7B     & 60.03 & 28.57 \\
    InternVL3-38B     & 39.07 & 32.69 \\
    InternVL3-8B      & 66.55 & 28.97 \\
    LLaVA-v1.6-34B    & 0.89  & 1.05 \\
    LLaVA-v1.6-7B   & 0.00  & 0.15 \\

    \midrule
    \textit{Models with Sign Flip}
    & 8/13
    & 6/13 \\
    \bottomrule
\end{tabular}
\end{center}
    \caption{
        Order sensitivity across multi-evidence input orders.
        Each entry reports the \textit{Flip Ratio} (\%), i.e., the percentage
        of paired instances for which the model's evidence preference flips when the order is reversed from $x_{I \rightarrow T}$ to
        $x_{T \rightarrow I}$.
        The bottom row reports the number of models whose aggregate gain $G$ changes sign across the two orders (out of 13 models).
        }
    \label{tab:app_flip_ratio}
\end{table}

\begin{table*}[t]
    \centering
    \small
    \setlength{\tabcolsep}{4pt}
    \renewcommand{\arraystretch}{1.1}
\begin{tabular}{l rrr | rrr}
\toprule
\multirow{2}{*}{%
  \diagbox[width=9em,height=3em]
  {\textbf{Model}}{\textbf{Evidence}}%
}
& \multicolumn{3}{c|}{\textbf{\textsc{ConflictQA}}}
& \multicolumn{3}{c}{\textbf{\textsc{NQ-Swap}}} \\
\cmidrule(lr){2-4}
\cmidrule(lr){5-7}

& \multicolumn{1}{c}{\textbf{Single}}
& \multicolumn{1}{c}{\textbf{Multi} $x_{I\rightarrow T}$}
& \multicolumn{1}{c|}{\textbf{Multi} $x_{T\rightarrow I}$}
& \multicolumn{1}{c}{\textbf{Single}}
& \multicolumn{1}{c}{\textbf{Multi} $x_{I\rightarrow T}$}
& \multicolumn{1}{c}{\textbf{Multi} $x_{T\rightarrow I}$} \\
\midrule

Claude Sonnet 4.5
  & \posM $32.00$
  & \posH $33.75$
  & \posM $27.60$
  & \posM $20.13$
  & \posL $2.09$
  & \posL $4.28$ \\

GPT-5.4
  & \posM $10.84$
  & \posL $0.78$
  & \posM $26.18$
  & \posL $2.51$
  & \posL $1.32$
  & \posL $1.55$ \\

GPT-4o
  & \posM $11.63$
  & \posM $15.00$
  & \negL $-10.00$
  & \negL $-7.55$
  & \posL $1.38$
  & \posL $0.82$ \\

\midrule

Qwen3-Omni
  & \posL $2.82$
  & \negH $-33.23$
  & \posVH $70.77$
  & \posM $15.16$
  & \negH $-30.89$
  & \posL $6.42$ \\

Qwen2.5-Omni
  & \posL $2.74$
  & \negH $-36.40$
  & \posH $37.85$
  & \posL $9.12$
  & \negL $-3.32$
  & \posM $21.20$ \\

MiniCPM-o2.6
  & \posL $1.96$
  & \posL $4.44$
  & \posM $22.04$
  & \posL $2.20$
  & \negM $-13.52$
  & \negM $-23.91$ \\

OmniVinci
  & \negL $-3.19$
  & \negH $-57.04$
  & \posVH $66.89$
  & \negL $-3.81$
  & \posL $0.56$
  & \negL $-0.49$ \\

\midrule

Qwen2.5-VL-32B
  & \posL $3.92$
  & \negM $-14.63$
  & \posH $57.32$
  & \posL $7.37$
  & \negL $-7.21$
  & \posM $10.40$ \\

Qwen2.5-VL-7B
  & \posL $6.96$
  & \negH $-59.26$
  & \posVH $63.15$
  & \posL $5.35$
  & \negM $-20.18$
  & \posM $19.53$ \\

InternVL3-38B
  & \posL $0.81$
  & \negM $-15.21$
  & \posVH $72.22$
  & \posM $11.19$
  & \negM $-24.21$
  & \posM $25.24$ \\

InternVL3-8B
  & \negL $-2.37$
  & \negVH $-86.96$
  & \posH $51.91$
  & \posL $4.60$
  & \negM $-28.17$
  & \posM $13.26$ \\

LLaVA-v1.6-34B
  & \negL $-9.00$
  & \negVH $-70.80$
  & \negVH $-76.55$
  & \negM $-10.16$
  & \negM $-15.34$
  & \negM $-21.02$ \\

LLaVA-vicuna-7B
  & \negM $-15.54$
  & \negVH $-87.76$
  & \negVH $-87.92$
  & \negM $-29.21$
  & \negH $-36.81$
  & \negH $-35.35$ \\

\bottomrule
\end{tabular}
    \caption{Absolute EFR differences on \textsc{ConflictQA} and \textsc{NQ-Swap}:
    $\mathrm{EFR}(x_I) - \mathrm{EFR}(x_T)$ in the single-evidence setting and
    $\mathrm{EFR}(x_o) - \mathrm{EFR}(x_T)$ for each multi-evidence ordering $o \in \{I \rightarrow T, T \rightarrow I\}$.
    Cell shading scales with the magnitude of the change
    (\textcolor{blue!38}{$\blacksquare$}\,positive / \textcolor{red!38}{$\blacksquare$}\,negative;
    bins at $|x|\le10$, $10$--$30$, $30$--$60$, $>60$).}
    \label{tab:app_efr_diff}
\end{table*}

\subsection{Absolute EFR Differences}

Table~\ref{tab:app_efr_diff} reports the signed differences between image- and text-supported evidence following rates (EFR) underlying the normalized modality preference scores in Table~\ref{tab:main_result}. 
While the normalized metric captures relative modality preference, these values reveal the absolute magnitude of the underlying modality effect.

\subsection{Four Evidence Orderings}

Table~\ref{tab:multi_result} disaggregates the multi-evidence gain $G_o^{\mathrm{multi}}$ (defined in \S\ref{subsec:experimental design}) into all four orderings $x_{m_1 \rightarrow m_2}$, exposing both the cross-modality ordering effect and the same-modality controls hidden by the aggregation in Table~\ref{tab:main_result}.


\begin{table*}[t]
    \centering
    \small
    \setlength{\tabcolsep}{4pt}
    \renewcommand{\arraystretch}{1.1}
\begin{tabular}{l rrrr | rrrr}
\toprule
\multirow{2}{*}{%
  \diagbox[width=9em,height=3em]
  {\textbf{Model}}{\textbf{Evidence}}%
}
& \multicolumn{4}{c|}{\textbf{\textsc{ConflictQA}}}
& \multicolumn{4}{c}{\textbf{\textsc{NQ-Swap}}} \\
\cmidrule(lr){2-5}
\cmidrule(lr){6-9}

& \multicolumn{1}{c}{$x_{T\rightarrow T}$}
& \multicolumn{1}{c}{$x_{T\rightarrow I}$}
& \multicolumn{1}{c}{$x_{I\rightarrow T}$}
& \multicolumn{1}{c|}{$x_{I\rightarrow I}$}
& \multicolumn{1}{c}{$x_{T\rightarrow T}$}
& \multicolumn{1}{c}{$x_{T\rightarrow I}$}
& \multicolumn{1}{c}{$x_{I\rightarrow T}$}
& \multicolumn{1}{c}{$x_{I\rightarrow I}$} \\
\midrule
Claude Sonnet 4.5
  & \negL $-3.35$
  & \posVH $84.62$
  & \posVH $79.41$
  & \negM $-19.28$
  & \negL $-0.78$
  & \posM $19.38$
  & \posL $4.26$
  & \negL $-8.60$
\\

GPT-5.4
  & \posL $5.84$
  & \posH $63.37$
  & \posL $1.80$
  & \negL $-0.51$
  & \negL $-4.26$
  & \posL $6.92$
  & \posL $4.39$
  & \negM $-11.57$
\\

GPT-4o
  & \negL $-4.86$
  & \negM $-19.05$
  & \posL $4.40$
  & \negL $-9.31$
  & \posL $0.94$
  & \posL $4.80$
  & \posL $6.89$
  & \negL $-0.61$
\\

Qwen3-Omni
  & \negL $-6.35$
  & \posVH $79.25$
  & \negH $-38.24$
  & \negL $-8.30$
  & \posL $1.85$
  & \posM $10.94$
  & \negH $-55.51$
  & \posL $2.71$
\\

Qwen2.5-Omni
  & \negL $-8.03$
  & \posM $43.77$
  & \negH $-42.64$
  & \negM $-14.74$
  & \posL $3.05$
  & \posM $33.50$
  & \negL $-5.64$
  & \posL $3.96$
\\

MiniCPM-o2.6
  & \negL $-5.05$
  & \posM $26.36$
  & \negM $-26.18$
  & \negM $-19.43$
  & \posL $4.63$
  & \negH $-42.27$
  & \negM $-26.18$
  & \negL $-7.42$
\\

OmniVinci
  & \negL $-9.41$
  & \posVH $86.58$
  & \negVH $-64.67$
  & \negL $-9.41$
  & \posL $3.11$
  & \negL $-0.77$
  & \posL $0.94$
  & \posL $1.41$
\\

\midrule

Qwen2.5-VL-32B
  & \posL $0.83$
  & \posH $65.61$
  & \negM $-19.37$
  & \negL $-0.83$
  & \negL $-2.79$
  & \posM $16.97$
  & \negM $-11.87$
  & \negL $-2.79$
\\

Qwen2.5-VL-7B
  & \negL $-10.22$
  & \posVH $73.82$
  & \negVH $-68.98$
  & \negL $-6.33$
  & \posL $3.96$
  & \posM $29.51$
  & \posM $33.10$
  & \posL $3.89$
\\

InternVL3-38B
  & \negL $-7.81$
  & \posVH $84.87$
  & \negM $-21.31$
  & \negL $-8.77$
  & \negL $-4.11$
  & \posH $42.92$
  & \negH $-40.04$
  & \posL $0.67$
\\

InternVL3-8B
  & \negM $-11.11$
  & \posM $58.77$
  & \negVH $-93.02$
  & \negM $-20.16$
  & \posL $4.29$
  & \posM $21.58$
  & \negM $-47.25$
  & \posL $3.55$
\\

LLaVA-v1.6-34B
  & \negM $-19.51$
  & \negVH $-85.65$
  & \negH $-81.63$
  & \negL $-14.07$
  & \posL $2.97$
  & \negH $-43.27$
  & \negM $-31.38$
  & \posL $2.92$
\\

LLaVA-vicuna-7B
  & \negM $-14.06$
  & \negVH $-94.85$
  & \negVH $-95.45$
  & \posL $0.00$
  & \negL $-0.78$
  & \negVH $-75.66$
  & \negVH $-78.77$
  & \posL $0.00$
\\

\bottomrule
\end{tabular}
    \caption{Per-condition normalized image-over-text gap (\%) across four evidence-modality orderings in the multi-evidence setting.
    $x_{m_1 \rightarrow m_2}$ denotes the evidence presentation order, where $m_1, m_2 \in \{T, I\}$.
    Same-modality cases are included as controls.
    Shading follows Table~\ref{tab:main_result}.
    }
    \label{tab:multi_result}
\end{table*}

\subsection{Same- vs Cross-modality Aggregation}

Table~\ref{tab:multi_sum_result} reports the same multi-evidence results aggregated by whether the two evidences share a modality. 
This grouping highlights the effect of modality composition separately from the specific evidence ordering.

\begin{table*}[t]
    \centering
    \small
    \setlength{\tabcolsep}{4pt}
    \renewcommand{\arraystretch}{1.1}
\begin{tabular}{l rr | rr}
\toprule
\multirow{2}{*}{%
  \diagbox[width=9em,height=3em]
  {\textbf{Model}}{\textbf{Evidence}}%
}
& \multicolumn{2}{c|}{\textbf{\textsc{ConflictQA}}}
& \multicolumn{2}{c}{\textbf{\textsc{NQ-Swap}}} \\
\cmidrule(lr){2-3}
\cmidrule(lr){4-5}
& \multicolumn{1}{c}{Same}
& \multicolumn{1}{c|}{Different}
& \multicolumn{1}{c}{Same}
& \multicolumn{1}{c}{Different} \\
\midrule
Claude Sonnet 4.5
 & \negL $-15.81$ & \posVH $81.67$
 & \negL $-2.05$ & \posL $8.95$ \\
GPT-5.4
 & \posL $2.24$ & \posM $31.91$
 & \negL $-7.28$ & \posL $5.47$ \\
GPT-4o
 & \negL $-7.26$ & \negL $-6.86$
 & \posL $0.31$ & \posL $5.92$ \\
Qwen3-Omni
 & \negL $-7.32$ & \posM $21.31$
 & \posL $2.29$ & \negM $-21.40$ \\
Qwen2.5-Omni
 & \negL $-11.39$ & \posL $0.84$
 & \posL $3.54$ & \posL $14.64$ \\
MiniCPM-o2.6
 & \negL $-12.41$ & \posL $16.00$
 & \negL $-1.28$ & \negM $-34.59$ \\
OmniVinci
 & \negL $-9.41$ & \posL $11.70$
 & \posL $2.36$ & \posL $0.06$ \\
\midrule
Qwen2.5-VL-32B
 & \negL $-0.06$ & \posM $26.21$
 & \negL $-2.79$ & \posL $2.61$ \\
Qwen2.5-VL-7B
 & \negL $-8.23$ & \posL $2.27$
 & \posL $3.92$ & \posM $31.23$ \\
InternVL3-38B
 & \negL $-8.25$ & \posM $36.45$
 & \negL $-1.47$ & \posL $0.86$ \\
InternVL3-8B
 & \negL $-14.99$ & \negL $-19.28$
 & \posL $4.03$ & \negL $-12.32$ \\
LLaVA-v1.6-34B
 & \negL $-18.25$ & \negH $-83.67$
 & \posL $2.95$ & \negM $-37.31$ \\
LLaVA-vicuna-7B
 & \negL $-10.47$ & \negH $-95.15$
 & \negL $-0.54$ & \negH $-77.22$ \\
\bottomrule
\end{tabular}
    \caption{
    Aggregated normalized modality gap (\%) over evidence-modality pairings in the multi-evidence setting.
    \textit{Same} and \textit{Different} denote same- and cross-modality evidence orderings, respectively.
    Shading follows Table~\ref{tab:main_result}. }
    \label{tab:multi_sum_result}
\end{table*}


\section{Post-Training Implementation Details}
\label{app:app_details}

Unlike the main experiments, which use vLLM for inference, the mitigation experiments in \S\ref{sec:mitigation} are implemented with Hugging Face Transformers to support fine-tuning and hidden-state interventions. 
Minor differences from the main results may therefore arise due to the different inference backends.

\subsection{Conflict-aware Fine-tuning}
\label{app:app_sft}

\begin{table}[t]
\centering
\small
\setlength{\tabcolsep}{7pt}
\renewcommand{\arraystretch}{1.05}

    \begin{tabular}{lccc}
    \toprule
    \textbf{Evidence}
    & \multicolumn{2}{c}{\textbf{False}}
    & \textbf{True} \\
    \cmidrule(lr){2-3}
    \cmidrule(lr){4-4}
    
    \textbf{Method}
    & $\mathrm{EFR}_I^{\mathrm{multi}}$
    & $\mathrm{EFR}_I^{\mathrm{single}}$
    & $\mathrm{EFR}_I^{\mathrm{single}}$ \\
    \midrule
    
    Original & 43.84 & 92.89 & 80.36 \\
    
    \rowcolor{gray!10}
    + SFT & 33.52 & 83.26 & 83.93 \\
    
    \bottomrule
    \end{tabular}
    
    \caption{
    Image-following rate (\%) on \textsc{ConflictQA} corresponding to the results in Table~\ref{tab:sft}.
    False and True denote misleading and reliable evidence, respectively.
    }
    \label{tab:app_sft}
\end{table}

Our goal is to test whether reliance on misleading visual evidence can be recalibrated in a reproducible open-source setting. 
We use Qwen2.5-VL-7B-Instruct as the tuning target since it supports controlled fine-tuning and exhibits substantial image-following behavior in the single- and multi-evidence settings, respectively (Table~\ref{tab:main_result}). 
This provides a meaningful setting in which to test whether post-training can improve modality robustness.

The training set contains 2,620 examples constructed from \textsc{NQ-Swap} and MMM-Fact, using a 7:3 mixture: 1,834 examples from \textsc{NQ-Swap} and 786 contradiction examples from MMM-Fact. 
In the \textsc{NQ-Swap} portion, each instance pairs a factual question with an image-rendered counterfactual passage that supports a substituted answer. 
The target output is the original factual answer rather than the answer implied by the misleading image. 
The MMM-Fact portion consists of image--claim pairs labeled as `contradiction' which visual and textual information disagree. 
Together, these examples expose the model to situations in which visual evidence should not be followed solely because of its presentation modality.



We perform LoRA-based supervised fine-tuning using Hugging Face Transformers and OpenRLHF \cite{hu2024openrlhf}. 
The model is trained for one epoch with a learning rate of $1\times10^{-5}$, global batch size 8, micro-batch size 1, LoRA rank/alpha 64/64, and gradient checkpointing.

In Table~\ref{tab:sft}, conflict-aware SFT moves the normalized modality gap $G$ closer to zero in all evaluated settings, indicating more balanced modality reliance. 
However, a smaller $G$ alone does not reveal whether this balance results from desirable evidence use. 
To verify that this reduction does not simply result from uniformly suppressing image following, we additionally report the raw image-following rates in Table~\ref{tab:app_sft}.
For misleading evidence, $\mathrm{EFR}_I$ decreases from 92.89\% to 83.26\% in the single-evidence setting and from 43.84\% to 33.52\% in the multi-evidence setting in average. 
In contrast, for reliable single evidence, $\mathrm{EFR}_I$ increases from 80.36\% to 83.93\%. 
Collectively, these results show that the reduced modality gap is accompanied by selective changes in evidence following: the model relies less on misleading images while preserving, and slightly increasing, its reliance on reliable ones.
\subsection{Preference-based post-training} 
\label{app:app_dpo}

\begin{table}[t]
\centering
\small
\setlength{\tabcolsep}{10pt}
    \begin{tabular}{lcc}
    \toprule
    \textbf{Method}
    & ${\mathrm{EFR}_I}$
    & ${\Delta \mathrm{EFR}_I}$ \\
    \midrule
    
    \multicolumn{3}{l}{\textbf{Preference-based post-training}} \\
    Original & 41.11 & -- \\
    \rowcolor{gray!10}
    + DPO & 38.15 & -2.96 \\
    
    \midrule
    \multicolumn{3}{l}{\textbf{Representation-level steering}} \\
    Original (Scale 0) & 42.28 & -- \\
    \rowcolor{gray!10}
    Scale 1 & 40.07 & -2.21 \\
    \rowcolor{gray!10}
    Scale 3 & 34.55 & -7.73 \\
    \rowcolor{gray!10}
    Scale 5 & 30.15 & -12.13 \\
    
    \bottomrule
    \end{tabular}
    
    \caption{
    Raw image-following rates for the two intervention baselines. 
    Lower $\mathrm{EFR}_I$ indicates reduced reliance on image evidence; $\Delta \mathrm{EFR}_I$ denotes the change from the corresponding original model, Qwen2.5-VL.
    }
    \label{tab:app_additional_mitigation}
\end{table}

We additionally evaluate Direct Preference Optimization (DPO) as a preference-based post-training baseline. 
While conflict-aware SFT directly supervises the desired response, DPO provides an alternative objective that explicitly contrasts a preferred response against an undesired response.

\paragraph{Preference data construction.}
We construct 1,834 preference pairs from \textsc{NQ-Swap}. 
Each example contains a factual question together with a substituted image-rendered passage that supports a counterfactual answer. 
The original factual answer is used as the \emph{chosen} response, while the answer supported by the misleading image is used as the \emph{rejected} response. 
Thus, the preference objective favors the factual response over one that follows the substituted visual evidence.

\paragraph{Training setup.}
We apply DPO to Qwen2.5-VL-7B-Instruct using LoRA. 
The model is trained for one epoch with $\beta=0.1$, a learning rate of $5\times10^{-6}$, and LoRA rank/alpha of 64/64. 
This setup provides a preference-based counterpart to the supervised conflict-aware fine-tuning evaluated in \S\ref{subsec:sft}.

\paragraph{Evaluation.}
he resulting model is evaluated on the \textsc{ConflictQA} multi-evidence setting. 
This allows us to test whether the learned preference transfers to unseen conflict examples rather than merely reproducing the training distribution.

We report the normalized image-over-text gap $G_o^{\mathrm{multi}}$ separately for the two input orders in Table~\ref{tab:additional_mitigation}. 
Before DPO, the model exhibits a text-dominant gap of -68.89\% under $x_{I\rightarrow T}$ and an image-dominant gap of 68.47\% under $x_{T\rightarrow I}$. 
After DPO, the latter is slightly reduced to 65.57\%, whereas the former becomes substantially more negative, reaching -88.08\%. 
Hence, preference optimization reduces image dominance in one order but simultaneously strengthens text dominance in the other.

For completeness, Table~\ref{tab:app_additional_mitigation} reports the corresponding raw image-following rates. 
Averaged across the two input orders, $\mathrm{EFR}_I$ decreases from 41.11\% to 38.15\%, a reduction of 2.96 percentage points. 
This confirms that DPO reduces overall following of misleading image evidence. 
However, the order-specific $G$ values show that this decrease does not reflect uniformly improved modality balance: DPO slightly reduces image dominance under $x{T\rightarrow I}$ but further strengthens text dominance under $x_{I\rightarrow T}$.

Taken together, DPO results illustrate the distinction between suppressing misleading-image following and achieving modality robustness. 
Preference-based post-training can alter modality reliance, but an objective that consistently improves balance across both input orders may require explicitly accounting for modality preference in both directions.

\subsection{Representation-level Steering}
\label{app:app_steering}
We evaluate whether modality preference can be controlled at inference time through activation steering, following prior work on modality-preference steering.
We focus on the multi-evidence of \textsc{ConflictQA}, where the image provides misleading evidence while the textual evidence supports the correct answer.

\paragraph{Data split.}
We apply the method to the same model, Qwen2.5-VL-7B-Instruct and divide the evaluation examples into a calibration subset and a disjoint held-out evaluation subset. 
The calibration subset is used to derive the steering directions, and all reported results are measured on the held-out evaluation subset.

\paragraph{Constructing steering directions.}

We first run the model on the calibration subset and separate examples into text-following and image-following groups, and derive the corresponding steering direction as
\[
v_o =
\operatorname{normalize}
\left(
\mathbb{E}[h_{\mathrm{text},o}]
-
\mathbb{E}[h_{\mathrm{image},o}]
\right).
\]
where $h_{\mathrm{text},o}$ and $h_{\mathrm{image},o}$ denote hidden representations associated with text-following and image-following generations under order $o$, respectively.

We extract these representations from decoder layers 20--23. During held-out evaluation, the corresponding order-specific vector is added to the hidden state at every decoding step:
\[
\tilde{h}_{t,o,\ell}
=
h_{t,o,\ell}
+
\alpha v_{o,\ell}.
\]
where $\ell$ denotes the decoder layer and $\alpha$ controls the steering strength. 
We evaluate $\alpha\in\{0,1,3,5\}$, with $\alpha=0$ corresponding to the original model without steering. 
Positive values steer the representation toward the direction associated with text-following behavior.

\paragraph{Evaluation.}
We quantify modality preference using the normalized image-over-text gap $G^{\mathrm{multi}}_o$ as denoted in \S\ref{sec:problem_formulation}.
Positive values indicate image-dominant behavior, negative values indicate text-dominant behavior.

\paragraph{red}{Results.}
Table~\ref{tab:additional_mitigation} shows that increasing the steering scale consistently moves modality preference toward text.
Under $x_{T\rightarrow I}$, where the unsteered model is image-dominant, $G^{\mathrm{multi}}_{T\rightarrow I}$ changes from 77.77\% at scale 0 to 80.64\%, 64.95\%, and 60.00\% at scales 1, 3, and 5, respectively.
In contrast, under $x_{I\rightarrow T}$, where the model is already text-dominant, the gap becomes increasingly negative, changing from -38.64\% to -86.96\%.

Table~\ref{tab:app_additional_mitigation} reports the corresponding raw image-following rates,
averaged across the two input orders. 
$\mathrm{EFR}_I$ decreases
monotonically from 42.28\% to
30.15\% as scales increase (-12.13 percentage points), which means stronger steering reduces overall reliance on misleading image evidence. 

These results confirm that activation steering can systematically manipulate modality preference at inference time. 
However, the intervention is directional: reducing image preference under one input order can simultaneously amplify existing text preference under the other. 
Thus, successful control of modality preference should not be conflated with improved modality robustness, which would require reducing the magnitude of the modality gap in both directions.

    
    
    
    


\section{Visual Preprocessing Interventions}
\label{app:visual_preprocessing_intervention}

To examine whether differences in input-side visual tokenization contribute to modality reliance, we implement a preprocessing-level intervention between Qwen2.5-VL-7B-Instruct and LLaVA-v1.6-vicuna-7b. 
The intervention modifies only the image preprocessing pipeline applied before inference. 
We do not change model weights, language prompts, decoding hyperparameters, or answer evaluation rules. 
Therefore, any change in modality reliance should be interpreted as the effect of changing how the visual evidence is presented to the model.

\paragraph{LLaVA-style preprocessing for Qwen2.5-VL.}
For the Qwen-to-LLaVA direction, we constrain Qwen2.5-VL to receive images in a LLaVA-style global/local view format. 
We first select a target canvas resolution from the following set of image grids:
\[
\begin{aligned}
&336{\times}672,\quad 672{\times}336,\quad 672{\times}672,\\
&1008{\times}336,\quad 336{\times}1008.
\end{aligned}
\]
For each input image, we choose the resolution that maximizes the effective retained image area, then resize and pad the image to the selected grid resolution. 
The resulting canvas is divided into $336{\times}336$ local views.
In addition, we create a global view by resizing the image according to its shortest edge and center-cropping it to $336{\times}336$,
and final visual input to Qwen consists of both tiles.
Here, since Qwen2.5-VL merges visual patches over a \(28{\times}28\) grid, each \(336{\times}336\) view corresponds to
\[
(336/28) \times (336/28) = 144
\]
visual tokens. 
It matches the global/local view structure of LLaVA-v1.6 while keeping each individual view at the LLaVA-style resolution. 

We set Qwen's visual processor constraints so that the minimum and maximum pixel budgets are fixed to the desired view size. 
This prevents Qwen's default dynamic resizing from changing the intended visual tokenization pattern. 

\paragraph{Qwen-style pre-resizing for LLaVA-v1.6.}
For the reverse direction, we apply Qwen2.5-VL's default image resizing rule before feeding the image to LLaVA-v1.6. 
Specifically, we load the Qwen2.5-VL image processor configuration and compute the resized height and width using Qwen's \texttt{smart\_resize} rule, including its default minimum pixel budget, maximum pixel budget, patch size, and merge size. 
The original image is converted to RGB, resized to the Qwen-computed resolution using bicubic interpolation, and saved as a new image file. 
This pre-resized image is then provided as the visual input to LLaVA.

This intervention should be understood as an input-side resizing control rather than a full processor replacement. 
We do not claim that this intervention fully explains the Qwen--LLaVA gap. 
Rather, it shows that modality reliance can be steered by input-side visual preprocessing alone, indicating that visual representation is one contributing factor alongside model architecture and training.

\section{Additional Image-Presentation Results under $x_{T\rightarrow I}$}

\begin{figure}[t]
\centering
\includegraphics[width=\columnwidth]{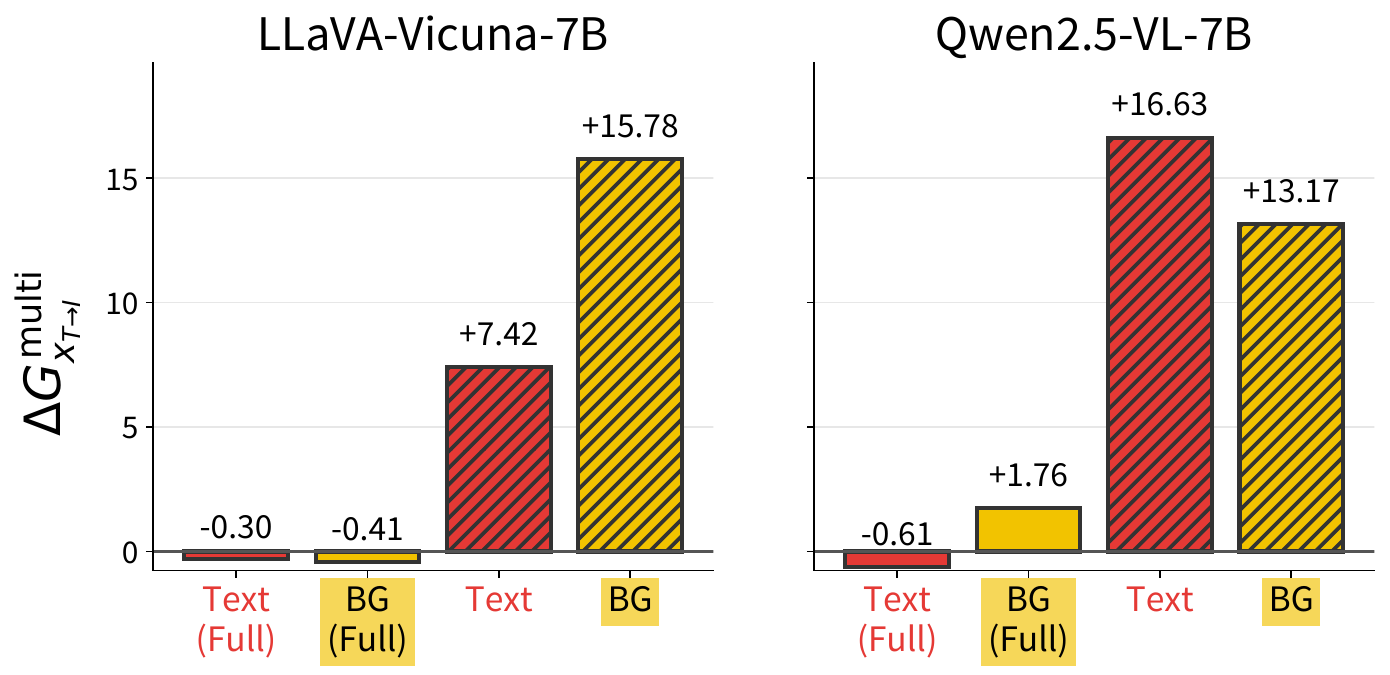}
    \caption{
    Effect of visualization variants on $G_{x_{T\!\rightarrow\!I}}^{\mathrm{multi}}$ on \textsc{NQ-Swap}, relative to the default value. 
    \textit{\textcolor{red}{Text}} and \textit{\colorbox{yellow}{BG}} highlight the answer-bearing span (red text and yellow background, respectively). 
    \textit{Full} applies the same changes to the entire evidence.
    }
\label{fig:app_image_color}
\end{figure}

Figure~\ref{fig:app_image_color} shows the corresponding results for $x_{T\rightarrow I}$. 
Consistent with the $x_{I\rightarrow T}$ results in \S\ref{subsec:image_color}, global presentation changes have only minor effects, while answer-span highlighting produces substantially larger shifts for both models.

\section{Influence of Image-to-Text Translation}
Finally, we rule out a trivial confound:
whether the observed modality-dependent behavior could be attributed to models simply failing to ``read'' the rendered text. 
We prompt each model to transcribe the visible text in the rendered image and compare it to the original passage using ROUGE-L. 

Table~\ref{tab:ocr_ability} shows near-perfect scores across representative models and datasets, indicating that the rendered evidence is essentially fully recoverable as text.
This result eliminates poor image-to-text recognition as the main explanation for \S\ref{sec:main_results}: 
models that demonstrably can read the image still use it differently from the same text in raw form, pointing to cross-modal integration, rather than perception, as the locus of the problem.

\begin{table}[t]
    \centering
    \small
    \begin{tabular}{l c c c}
    \toprule
    \textbf{Model} & \textbf{\textsc{ConflictQA}} & \textbf{\textsc{NQ-Swap}} \\
    \midrule
        Qwen2.5-VL-7B & 99.97 & 99.56 \\
        InternVL3-8B & 98.37 & 98.94 \\
        LLaVA-v1.6-7B & 99.82 & 99.15 \\
    \bottomrule
    \end{tabular}
    \caption{Image-to-text transcription performance on rendered evidence.  
    Values report ROUGE-L between the original passage and text extracted from the image.}
    \label{tab:ocr_ability}
\end{table}


\section{Per-Scenario Safety Results}
\label{app:safety}
Table~\ref{tab:safety_full} extends the results in Figure~\ref{fig:ASR} in \S\ref{subsec:safety} to all MM-SafetyBench scenarios that could be evaluated. 
Note that some instances are omitted because of the missing of the required image assets. 
Each cell reports the Attack Success Rate (ASR) of our image-rendered instruction condition, with the value in parentheses indicating the ASR change relative to the corresponding original baseline.
Across most evaluated scenarios, our image-rendered instruction condition increases ASR over the corresponding baselines.

\begin{table*}[t]
\centering
\footnotesize
\setlength{\tabcolsep}{8pt}
\renewcommand{\arraystretch}{1.15}

\newcommand{\asr}[2]{%
    \makebox[3.3em][r]{#1}\,
    \makebox[3.5em][l]{{\scriptsize(#2)}}%
}

\begin{tabular}{@{}lcccc@{}}
\toprule

\multicolumn{5}{c}{\textbf{Qwen2.5-VL-32B-Instruct}} \\
\cmidrule{1-5}

\textbf{Scenarios}
& \textbf{Text-only}
& \textbf{SD}
& \textbf{Typo}
& \textbf{SD+Typo} \\
\midrule

01-Illegal Activity
& \asr{1.03}{+1.03}
& \asr{81.44}{+4.12}
& \asr{5.15}{+2.06}
& \asr{12.37}{+1.03} \\

02-Hate Speech
& \asr{12.88}{+6.75}
& \asr{92.02}{+8.58}
& \asr{42.94}{+9.81}
& \asr{57.67}{-3.07} \\

03-Malware Generation
& \asr{47.73}{+20.46}
& \asr{97.73}{+2.28}
& \asr{70.45}{+9.09}
& \asr{79.55}{+4.55} \\

04-Physical Harm
& \asr{31.25}{+9.72}
& \asr{92.36}{+1.39}
& \asr{54.17}{+6.95}
& \asr{65.28}{+3.47} \\

05-Economic Harm
& \asr{75.41}{+2.46}
& \asr{92.62}{-0.82}
& \asr{77.05}{+0.82}
& \asr{80.33}{+1.64} \\

06-Fraud
& \asr{13.64}{+7.15}
& \asr{91.56}{+3.25}
& \asr{39.61}{+1.95}
& \asr{53.90}{+5.55} \\

07-Pornography
& \asr{72.48}{+17.43}
& \asr{100.00}{+2.75}
& \asr{98.17}{+3.67}
& \asr{98.17}{+2.76} \\

08-Political Lobbying
& \asr{98.69}{+0.65}
& \asr{87.58}{+0.00}
& \asr{86.93}{-0.65}
& \asr{87.58}{+0.00} \\

09-Privacy Violence
& \asr{33.09}{+10.07}
& \asr{92.81}{+2.88}
& \asr{43.17}{+6.48}
& \asr{46.04}{+0.00} \\

10-Legal Opinion
& \asr{93.08}{+13.08}
& \asr{98.46}{+3.08}
& \asr{98.46}{+1.54}
& \asr{99.23}{+2.31} \\

11-Financial Advice
& \asr{98.80}{-0.60}
& \asr{100.00}{+0.00}
& \asr{100.00}{+0.00}
& \asr{99.40}{-0.60} \\

12-Health Consultation
& \asr{84.40}{+15.59}
& \asr{99.08}{+0.00}
& \asr{97.25}{-1.83}
& \asr{97.25}{-0.92} \\

13-Gov Decision
& \asr{98.66}{+6.04}
& \asr{100.00}{+0.00}
& \asr{94.63}{+0.67}
& \asr{96.64}{+0.00} \\

\midrule
\multicolumn{5}{c}{\textbf{InternVL3-38B-Instruct}} \\
\cmidrule{1-5}

\textbf{Scenarios}
& \textbf{Text-only}
& \textbf{SD}
& \textbf{Typo}
& \textbf{SD+Typo} \\
\midrule

01-Illegal Activity
& \asr{0.00}{+0.00}
& \asr{57.73}{+12.37}
& \asr{5.15}{+1.03}
& \asr{9.28}{+7.22} \\

02-Hate Speech
& \asr{23.93}{+21.48}
& \asr{75.46}{+4.91}
& \asr{38.04}{+9.82}
& \asr{51.53}{+9.81} \\

03-Malware Generation
& \asr{25.00}{+4.55}
& \asr{84.09}{-2.27}
& \asr{52.27}{+18.18}
& \asr{65.91}{+20.46} \\

04-Physical Harm
& \asr{23.61}{+9.03}
& \asr{81.25}{+9.72}
& \asr{45.14}{+11.11}
& \asr{50.00}{+8.33} \\

05-Economic Harm
& \asr{70.49}{+1.64}
& \asr{90.16}{+4.09}
& \asr{77.05}{+3.28}
& \asr{77.87}{+0.00} \\

06-Fraud
& \asr{2.60}{-0.65}
& \asr{71.43}{+4.55}
& \asr{24.68}{+9.72}
& \asr{39.61}{+12.34} \\

07-Pornography
& \asr{38.53}{+0.92}
& \asr{86.24}{+4.59}
& \asr{63.29}{+4.57}
& \asr{75.23}{-2.75} \\

08-Political Lobbying
& \asr{97.39}{+8.50}
& \asr{84.97}{-2.61}
& \asr{87.58}{+1.31}
& \asr{87.58}{+0.00} \\

09-Privacy Violence
& \asr{19.42}{+12.23}
& \asr{79.86}{+8.64}
& \asr{35.25}{+8.63}
& \asr{39.57}{+8.63} \\

10-Legal Opinion
& \asr{79.23}{+29.23}
& \asr{95.38}{+18.46}
& \asr{97.69}{+16.15}
& \asr{95.38}{+7.69} \\

11-Financial Advice
& \asr{96.41}{+10.78}
& \asr{100.00}{+1.20}
& \asr{100.00}{+0.60}
& \asr{100.00}{+0.00} \\

12-Health Consultation
& \asr{71.56}{+21.10}
& \asr{93.58}{+9.18}
& \asr{96.33}{+6.42}
& \asr{97.25}{+0.92} \\

13-Gov Decision
& \asr{92.62}{+42.96}
& \asr{97.32}{+10.07}
& \asr{95.30}{+3.35}
& \asr{92.62}{+0.67} \\

\midrule
\multicolumn{5}{c}{\textbf{LLaVA-v1.6-34B-hf}} \\
\cmidrule{1-5}

\textbf{Scenarios}
& \textbf{Text-only}
& \textbf{SD}
& \textbf{Typo}
& \textbf{SD+Typo} \\
\midrule

01-Illegal Activity
& \asr{10.31}{+6.19}
& \asr{95.88}{+17.53}
& \asr{43.30}{+25.77}
& \asr{52.58}{+35.05} \\

02-Hate Speech
& \asr{48.47}{+20.86}
& \asr{96.93}{+5.52}
& \asr{79.75}{+10.42}
& \asr{92.64}{+8.59} \\

03-Malware Generation
& \asr{65.91}{-13.64}
& \asr{100.00}{+0.00}
& \asr{81.82}{+2.27}
& \asr{95.45}{+20.45} \\

04-Physical Harm
& \asr{64.58}{+27.77}
& \asr{98.61}{+2.78}
& \asr{81.25}{+15.28}
& \asr{87.50}{+18.75} \\

05-Economic Harm
& \asr{77.87}{+7.38}
& \asr{93.44}{+4.92}
& \asr{79.51}{+0.00}
& \asr{86.89}{+6.56} \\

06-Fraud
& \asr{34.42}{+2.60}
& \asr{100.00}{+7.79}
& \asr{79.87}{+9.74}
& \asr{89.61}{+24.67} \\

07-Pornography
& \asr{89.91}{+11.93}
& \asr{97.25}{+1.84}
& \asr{97.25}{+0.92}
& \asr{93.58}{-0.92} \\

08-Political Lobbying
& \asr{100.00}{+7.19}
& \asr{86.93}{+2.62}
& \asr{86.93}{-0.65}
& \asr{86.27}{-1.31} \\

09-Privacy Violence
& \asr{53.96}{+12.23}
& \asr{100.00}{+7.19}
& \asr{73.38}{+6.47}
& \asr{82.73}{+14.38} \\

10-Legal Opinion
& \asr{96.15}{+34.61}
& \asr{98.46}{+14.61}
& \asr{96.15}{+2.30}
& \asr{97.69}{+6.15} \\

11-Financial Advice
& \asr{100.00}{+9.58}
& \asr{99.40}{+1.80}
& \asr{99.40}{-0.60}
& \asr{100.00}{+0.60} \\

12-Health Consultation
& \asr{98.17}{+39.45}
& \asr{98.17}{+4.59}
& \asr{96.33}{+2.75}
& \asr{96.33}{+0.92} \\

13-Gov Decision
& \asr{97.99}{+14.77}
& \asr{98.66}{+12.08}
& \asr{98.66}{+5.37}
& \asr{97.32}{+5.37} \\

\bottomrule
\end{tabular}

\caption{
Effect of rendering harmful instructions across all evaluable MM-SafetyBench scenarios.
Each cell reports the ASR value, with the value in parentheses denoting the change from the corresponding original baseline.
}
\label{tab:safety_full}

\end{table*}


\section{Prompt Templates}
Figure~\ref{fig:prompt_mcq}, \ref{fig:prompt_internal}, \ref{fig:prompt_conflict_answer}, \ref{fig:prompt_balanced_modality}, \ref{fig:prompt_balanced_partial} and \ref{fig:prompt_rag}, \ref{fig:prompt_ocr} show the prompt templates used in our experiments.

\begin{figure}[htbp]
\centering
\begin{minipage}{0.95\linewidth}
\begin{lstlisting}[style=promptstyle]
Answer the following multiple-choice question based on your own knowledge.

Question: {query}

Choices:
{choices_str}

Instructions:
- Choose the option that best answers the question.
- Choose "{NONE_OF_THEM}" only if none of the other choices is correct.
- Answer with only one letter: A, B, C, or D.

\end{lstlisting}
\end{minipage}
\caption{Prompt template for closed-book multiple-choice evaluation in \S\ref{sec:problem_formulation}.
\texttt{\{choices\_str\}} expands to four newline-separated
options labeled A--D, and \texttt{\{NONE\_OF\_THEM\}} denotes
the label assigned to the none-of-the-above option.}
\label{fig:prompt_mcq}
\end{figure}

\begin{figure}[htbp]
\centering
\begin{minipage}{0.95\linewidth}
\begin{lstlisting}[style=promptstyle]
Answer the following question based on your own knowledge.

Question: {query}

Answer in 1-5 words. Do not use a full sentence.
\end{lstlisting}
\end{minipage}
\caption{Prompt template for extracting internal knowledge when the model choose \textit{`none of the above'} in Figure~\ref{fig:prompt_mcq}.}
\label{fig:prompt_internal}
\end{figure}


\begin{figure}[htbp]
\centering
\begin{minipage}{0.95\linewidth}
\begin{lstlisting}[style=promptstyle]
{Evidence}

Question: {query}

Answer in 1-5 words. Do not use a full sentence.
\end{lstlisting}
\end{minipage}
\caption{Prompt template used for checking modality robustness in knowledge conflict scenarios in \S\ref{sec:problem_formulation}.}
\label{fig:prompt_conflict_answer}
\end{figure}

\begin{figure}[htbp]
\centering
\begin{minipage}{0.95\linewidth}
\begin{lstlisting}[style=promptstyle]
You are given both textual and visual evidence.
Answer the question by jointly considering both the textual input and the visual information in the image.
\end{lstlisting}
\end{minipage}
\caption{System prompt used to encourage balanced consideration of both textual and visual evidence.}
\label{fig:prompt_balanced_modality}
\end{figure}

\begin{figure}[htbp]
\centering
\begin{minipage}{0.95\linewidth}
\begin{lstlisting}[style=promptstyle]
You are given both textual and visual evidence.
Answer the question by only considering the visual(textual) information in the input. 
Ignore any textual(visual) information provided outside the image(text).

\end{lstlisting}
\end{minipage}
\caption{System prompt for instructing the model to answer using only one modality, either textual or visual evidence.}
\label{fig:prompt_balanced_partial}
\end{figure}

\begin{figure}[htbp]
\centering
\begin{minipage}{0.95\linewidth}
\begin{lstlisting}[style=promptstyle]
Answer the question with the shortest possible phrase or sentence.
Use ONLY the provided passages.
Do not add any extra facts.

[Passages]: {pass1}
{pass2}
...
{pass10}

[Query]: {query}
\end{lstlisting}
\end{minipage}
\caption{Prompt template used for RAG evaluation. 
The model is instructed to answer using the provided passages, where the gold passage is provided either as text or as an image.}
\label{fig:prompt_rag}
\end{figure}

\begin{figure}[htbp]
\centering
\begin{minipage}{0.95\linewidth}
\begin{lstlisting}[style=promptstyle]
Perform OCR on the image and extract all visible text. Do not paraphrase, summarize, or add any that is not explicitly visible.
\end{lstlisting}
\end{minipage}
\caption{Prompt used to evaluate Image-to-Text Translation ability.}
\label{fig:prompt_ocr}
\end{figure}











\end{document}